\documentclass[runningheads]{llncs}
\usepackage{graphicx}
\usepackage{amsmath,amssymb} 

\usepackage{subcaption}
\usepackage[]{caption}	
\usepackage{lscape}                                         
\usepackage[export]{adjustbox}

\usepackage[lined,ruled,linesnumbered]{algorithm2e}

\usepackage{booktabs}                   
\usepackage{multirow}

\usepackage{paralist}
\usepackage{enumitem}

\usepackage{bm}                          
\usepackage{epsfig}                      
\usepackage{times}
\usepackage{pifont}

\usepackage{units}
\usepackage{xcolor}

\usepackage{url}  
\usepackage{xspace}
\usepackage[percent]{overpic}
\usepackage{ifthen}

\def\etal{et al.~}			  
\def\ie{i.e.,~}               
\def\etc{etc}                 

\newlength\paramargin
\newlength\figmargin
\newlength\secmargin
\newlength\figcapmargin

\newcommand{\heading}[1]
{
\vspace{1mm}
\noindent \textbf{#1}
}   

\newcommand{\secref}[1]{Section~\ref{sec:#1}}
\newcommand{\figref}[1]{Figure~\ref{fig:#1}} 
\newcommand{\tabref}[1]{Table~\ref{tab:#1}}
\newcommand{\eqnref}[1]{Equation~\ref{eqn:#1}}

\long\def\ignorethis#1{}

\newcommand{\tb}[1]{\textbf{#1}}

\def\incompletedpano{{\mathbf{I}}_{in}}
\def\layout{{\mathbf{Layout\ map\ only}}}
\def\layoutthreeclass{{\mathbf{L}_{3-class}}}
\def\layoutplanewise{{\mathbf{L}_{p-wise}}}
\def\layoutgmap{{\mathbf{L}_{m}}}
\def\inpaintedpano{{\mathbf{I}}_{out}}
\def\groundtruthpano{{\mathbf{I}}_{gt}}
\def\mask{\mathbf{M}}

\def\generator{\mathbf{G}}

\def\totalloss{L_{total}}
\def\recloss{L_{rec}}
\def\oneloss{L_{1}}
\def\percloss{L_{perc}}
\def\styloss{L_{sty}}

\def\baseline{\mathbf{Backbone}}

\def\planarawarenor{\mathbf{Full\ model}}
\graphicspath{{Figures}, {images}, {example}}

\begin{document}
\title{Layout-guided Indoor Panorama Inpainting with Plane-aware Normalization}
\titlerunning{LGPN-net}
%
\author{Chao-Chen Gao\inst{1}\orcidID{0000-0002-5325-3592} \and
Cheng-Hsiu Chen\inst{1}\orcidID{0000-0002-8652-5830} \and
Jheng-Wei Su\inst{1}\orcidID{0000-0003-3148-002X} \and
Hung-Kuo Chu\inst{1}\orcidID{0000-0001-7153-4411}}
\authorrunning{Gao et al.}
%
\institute{$^1$National Tsing Hua University, Taiwan\\
\email{hkchu@cs.nthu.edu.tw}}

\maketitle              
%
\begin{figure}
   \begin{subfigure}[t]{.24\columnwidth}
       \includegraphics[width=\columnwidth,keepaspectratio]{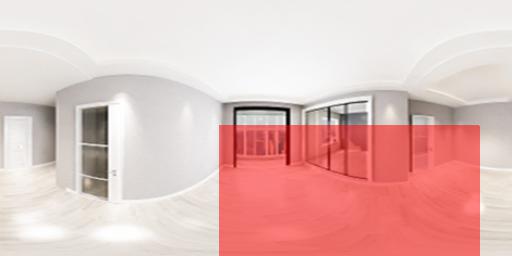}
       \includegraphics[width=\columnwidth,keepaspectratio]{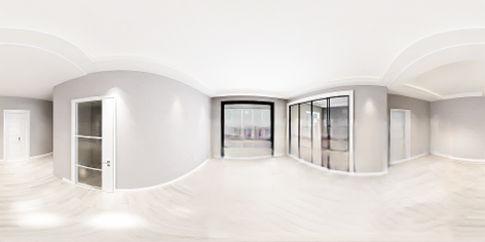}
              \caption{\resizebox{.85\textwidth}{!}{Synthetic empty scene}}
    \end{subfigure}
    \begin{subfigure}[t]{.24\columnwidth}
       \includegraphics[width=\columnwidth,keepaspectratio]{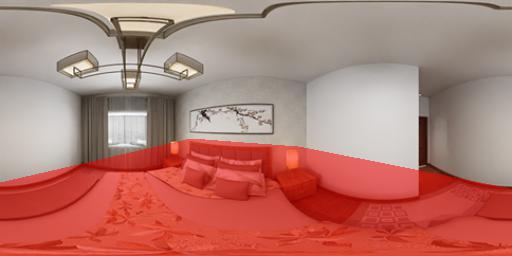}
       \includegraphics[width=\columnwidth,keepaspectratio]{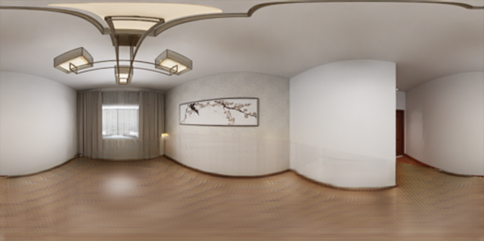}
              \caption{\resizebox{.85\textwidth}{!}{Synthetic furnished scene}}
    \end{subfigure}
    \begin{subfigure}[t]{.24\columnwidth}
       \includegraphics[width=\columnwidth,keepaspectratio]{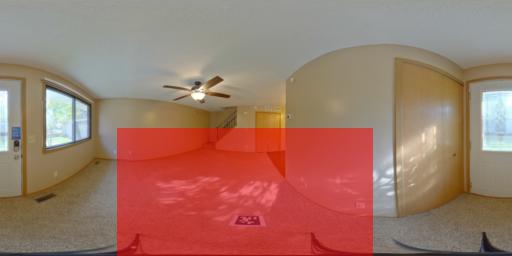}
       \includegraphics[width=\columnwidth,keepaspectratio]{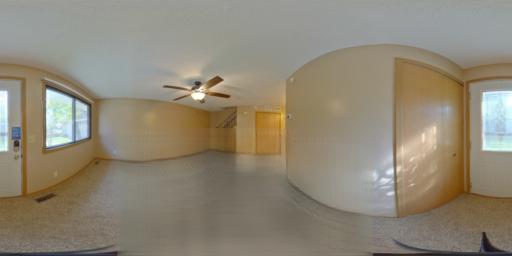}
          \caption{\resizebox{.85\textwidth}{!}{Real-world empty scene}}
    \end{subfigure}
    \begin{subfigure}[t]{.24\columnwidth}
       \includegraphics[width=\columnwidth,keepaspectratio]{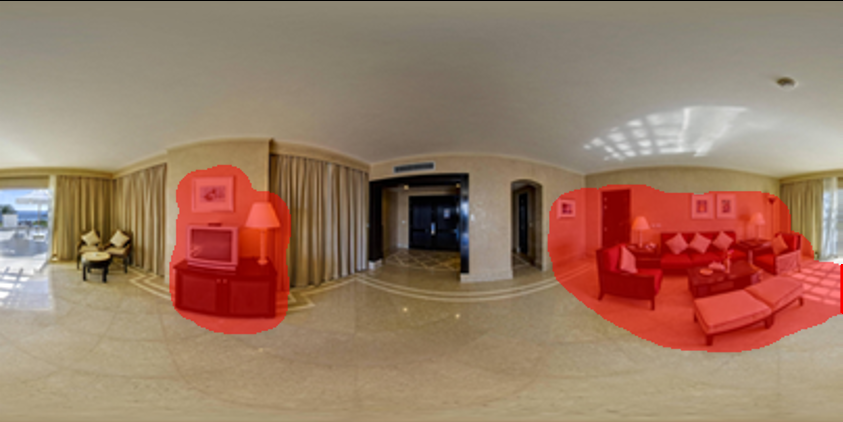}
       \includegraphics[width=\columnwidth,keepaspectratio]{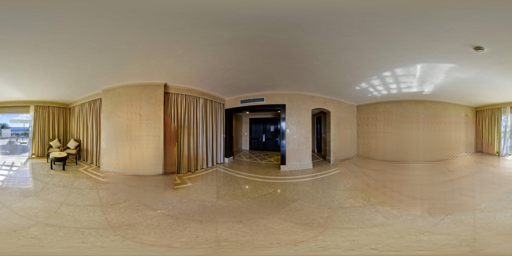}
          \caption{\resizebox{.85\textwidth}{!}{Real-world furnished scene}}
    \end{subfigure}
      \caption{\tb{Indoor panorama inpainting.} We present a learning-based indoor panorama inpainting method that is capable of generating plausible results for the tasks of hole filling (a)(c) and furniture removal (b)(d) in both synthetic (a)(b) and real-world (c)(d) scenes.  
    }
  \label{fig:teaser}
\end{figure}
%
\begin{abstract}
We present an end-to-end deep learning framework for indoor panoramic image inpainting. 
Although previous inpainting methods have shown impressive performance on natural perspective images, most fail to handle panoramic images, particularly indoor scenes, which usually contain complex structure and texture content. 
To achieve better inpainting quality, we propose to exploit both the global and local context of indoor panorama during the inpainting process. 
Specifically, we take the low-level layout edges estimated from the input panorama as a prior to guide the inpainting model for recovering the global indoor structure. 
A plane-aware normalization module is employed to embed plane-wise style features derived from the layout into the generator, encouraging local texture restoration from adjacent room structures (\ie ceiling, floor, and walls). 
Experimental results show that our work outperforms the current state-of-the-art methods on a public panoramic dataset in both qualitative and quantitative evaluations.
Our code is available online\footnote{\url{https://ericsujw.github.io/LGPN-net/}}.

\end{abstract}
\section{Introduction}
\label{sec:intro}
Image inpainting is a widely investigated topic in computer graphics and vision communities, which aims at filling in missing regions of an image with photorealistic and fine detailed content.
It plays a crucial step toward many practical applications, such as image restoration, object removal, \etc.
With the rapid development of deep learning, image inpainting has been revisited and improved significantly in the past few years.
A considerable body of researches has been explored to generate impressive results on perspective datasets. 

\begin{figure}[!t]
\centering
    \begin{subfigure}[t]{.24\columnwidth}
       \includegraphics[width=\columnwidth,keepaspectratio]{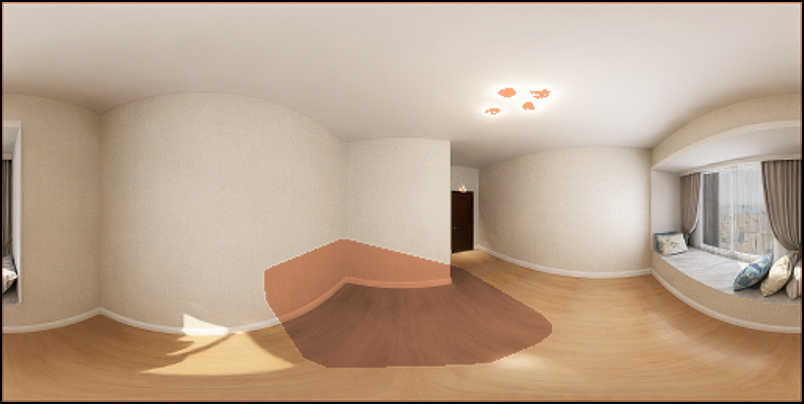}
       \includegraphics[width=\columnwidth,keepaspectratio]{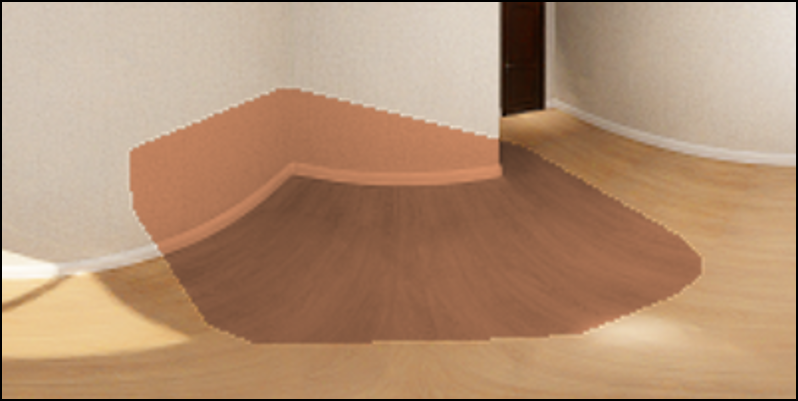}

       \includegraphics[width=\columnwidth,keepaspectratio]{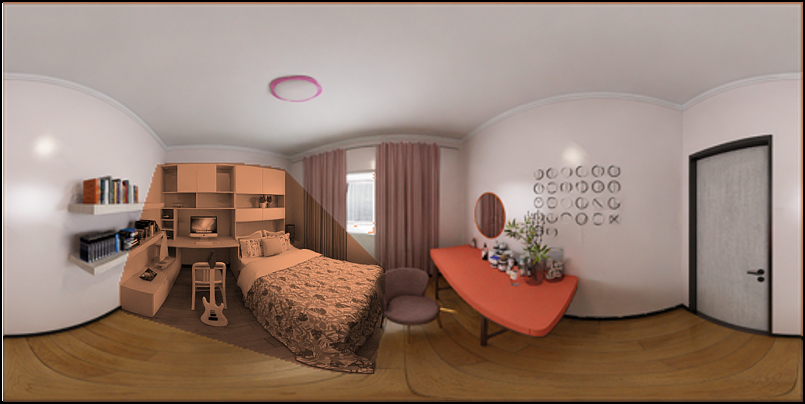}
       \includegraphics[width=\columnwidth,keepaspectratio]{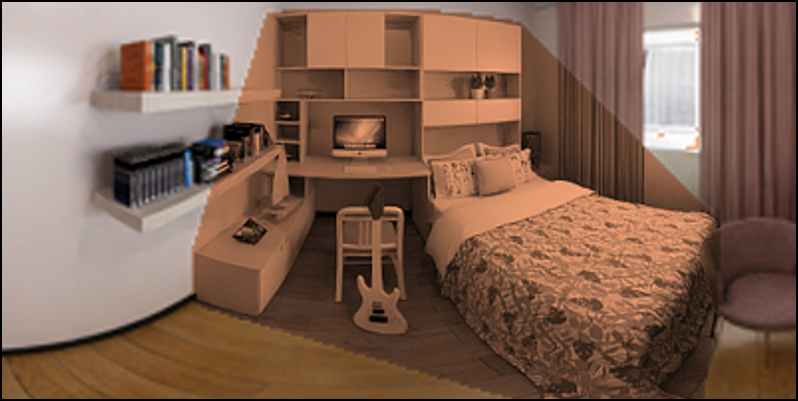}    
              \caption*{Input}
    \end{subfigure}
    \begin{subfigure}[t]{.24\columnwidth}
       \includegraphics[width=\columnwidth,keepaspectratio]{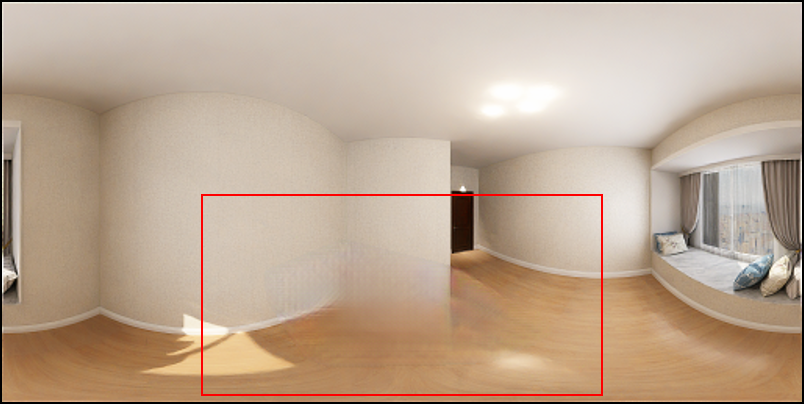}
       \includegraphics[width=\columnwidth,keepaspectratio]{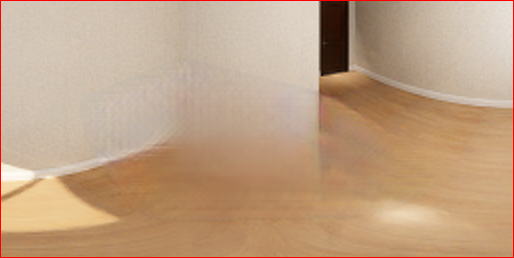}

       \includegraphics[width=\columnwidth,keepaspectratio]{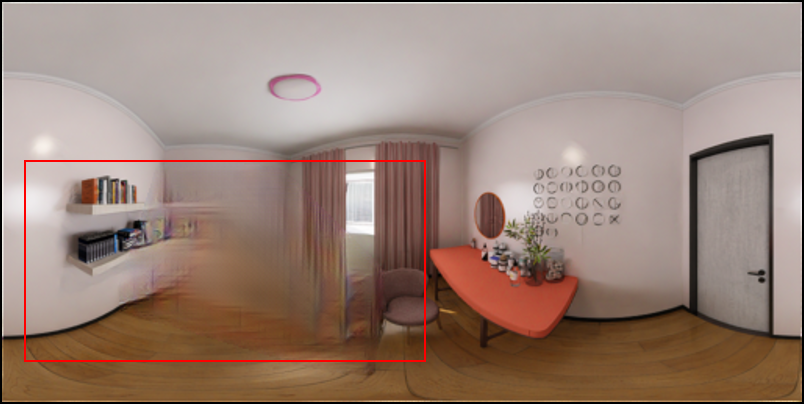}
       \includegraphics[width=\columnwidth,keepaspectratio]{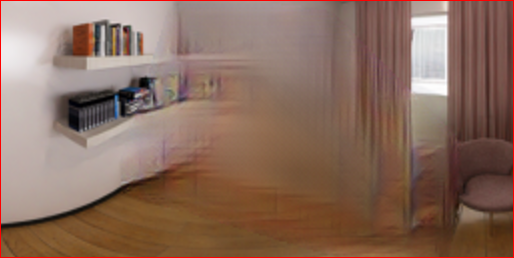}    
          \caption*{EC~\cite{nazeri2019edgeconnect}}
    \end{subfigure}
    \begin{subfigure}[t]{.24\columnwidth}
       \includegraphics[width=\columnwidth,keepaspectratio]{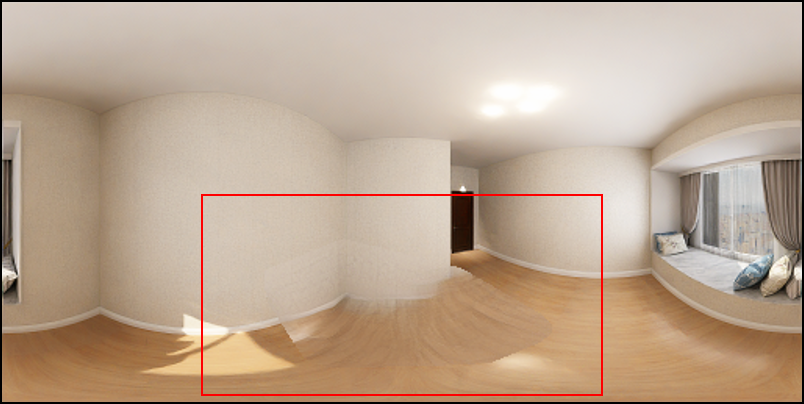}
       \includegraphics[width=\columnwidth,keepaspectratio]{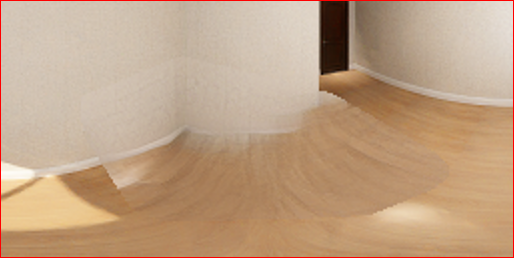}

       \includegraphics[width=\columnwidth,keepaspectratio]{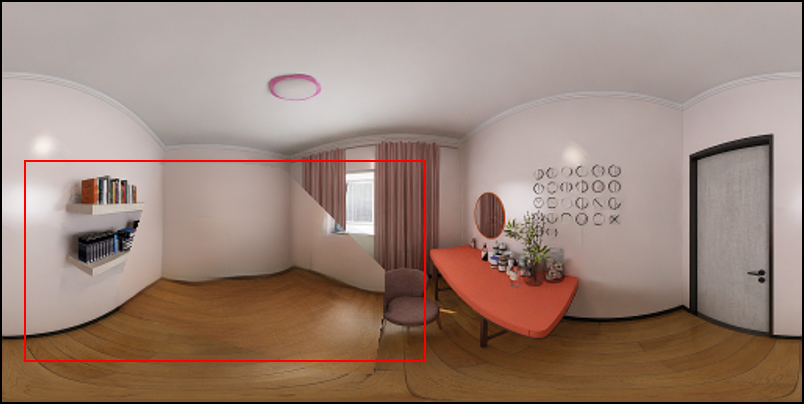}
       \includegraphics[width=\columnwidth,keepaspectratio]{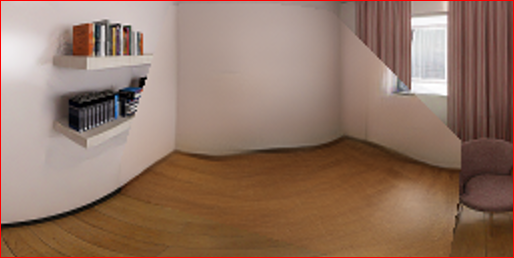}    
          \caption*{PanoDR~\cite{gkitsas2021panodr}}
    \end{subfigure}
    \begin{subfigure}[t]{.24\columnwidth}
       \includegraphics[width=\columnwidth,keepaspectratio]{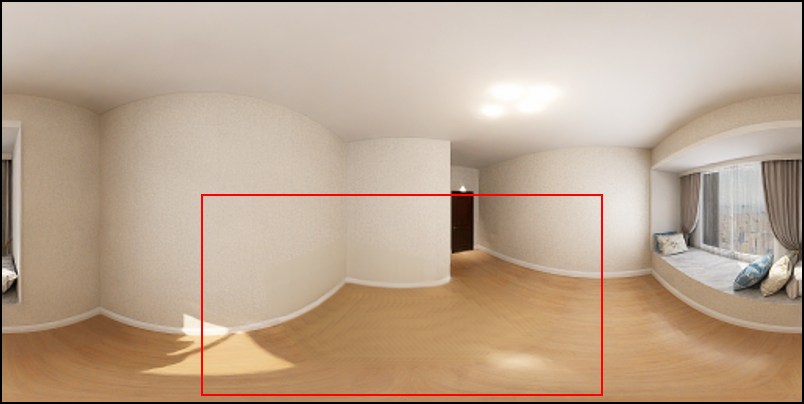}
       \includegraphics[width=\columnwidth,keepaspectratio]{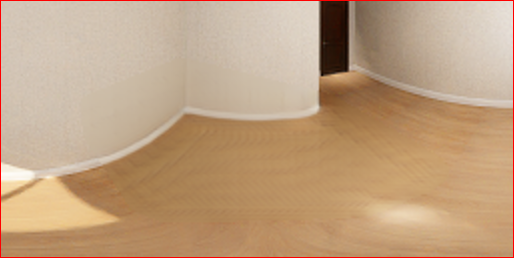}

       \includegraphics[width=\columnwidth,keepaspectratio]{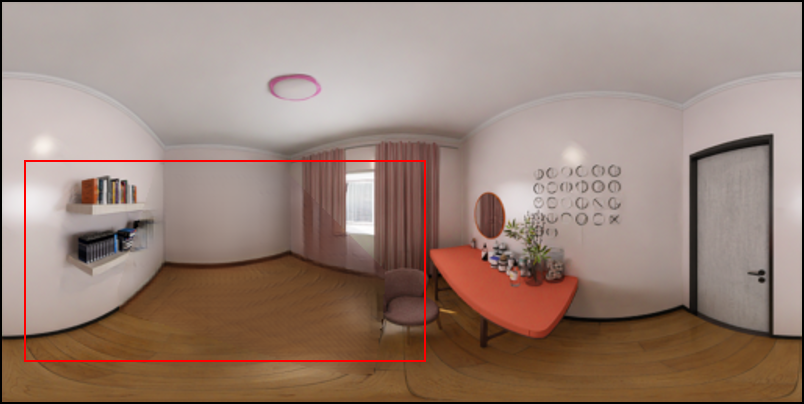}
       \includegraphics[width=\columnwidth,keepaspectratio]{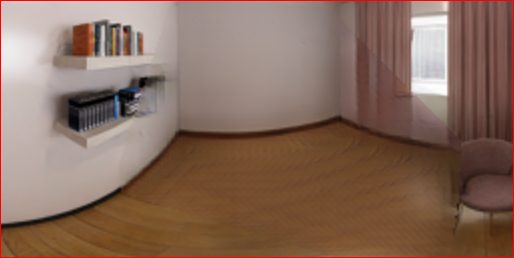}    
          \caption*{Ours}
    \end{subfigure}
    \caption{
    \tb{Limitations of existing methods.} 
    EC~\cite{nazeri2019edgeconnect} and PanoDR~\cite{gkitsas2021panodr} fail to align the image structure along the layout boundaries and produce inconsistent blurry image contents in the inpainted regions (red mask).}
    \label{fig:FR_SI}
\end{figure}

In this work, we address the image inpainting problem in the context of indoor panoramas.
Indoor panoramas provide excellent media for the holistic scene understanding~\cite{LNCS86940668panocontext} that would further benefit several applications such as object detection, depth estimation, furniture rearrangement, \etc.
In particular, removing foreground objects and filling the missing regions in an indoor panorama is essential for the interior redesign task.
However, the complex structures and textures presented in the indoor scenes make the inpainting problem non-trivial and challenging for previous methods.
As shown in~\figref{FR_SI}(EC), results generated by a state-of-the-art deep learning method fail to align the image structure along the layout boundaries and produce inconsistent blurry image contents. 

Recently, Gkitsas~\etal~\cite{gkitsas2021panodr} introduced PanoDR, a diminished reality-oriented inpainting model for indoor panorama.
The main idea is to translate a furnished indoor panorama into its empty counterpart via a network that leverages both a generator and an image-to-image translation module.
The inpainting result is then obtained by compositing the predicted empty panorama and input panorama using the object mask.
However, there are still obvious artifacts near the boundaries of masked regions as shown in~\figref{FR_SI}.

To achieve better inpainting quality, we present an end-to-end deep generative adversarial framework that exploits both the global and local context of indoor panoramas to guide the inpainting process.
Specifically, we take the low-level layout boundaries estimated from input panorama as a conditional input to guide the inpainting model, encouraging the preservation of sharp boundaries in the filled image.
A plane-aware normalization module is then employed to embed local plane-wise style features derived from the layout into the image decoder, encouraging local texture restoration from adjacent room structures (\ie ceiling, floor, and individual walls).
We train and evaluate our model on a public indoor panorama dataset, Structured3D~\cite{Structured3D}. 
Experimental results show that our method produces results superior to several state-of-the-art methods (see~\figref{teaser}, \figref{FR_SI} and \figref{comp_sota_si_fr}). The main contributions are summarized as follows:
\begin{itemize}
\item We present an end-to-end generative adversarial network that incorporates both the global and local context of indoor panoramas to guide the inpainting process.
\item We introduce a plane-aware normalization module that guides the image decoder with spatially varying normalization parameters per structural plane (\ie ceiling, floor, and individual walls).
\item Our method achieves state-of-the-art performance and visual quality on synthetic and real-world datasets. 
\end{itemize}
\section{Related Work}
\label{sec:related}

\heading{Traditional image inpainting.} 
There are two main genres among traditional image inpainting works: diffusion-based methods and patch-based methods.
Diffusion-based methods~\cite{mumford,10.1145/364338.364405,935036,Drori2003FragmentbasedIC,Liang2015AnEF,wei2016} propagate pixels from neighboring regions to the missing ones to synthesize the image content.
On the other hand, patch-based methods~\cite{patchmatch,6960838,7180400,7922581,10.1007/s11042-017-4509-0,8352516,10.1007/s11042-017-5077-z} fill the missing regions by searching for and copying similar image patches from the rest of the image or existing image datasets.
Without a high-level understanding of the image contents, these methods easily fail on images with complex structures.

\heading{Learning-based image inpainting.} 
With the rapid development of deep learning, several image inpainting techniques based on convolutional neural networks (CNN) have been proposed.
These methods aim to learn a generator from a large dataset to produce photorealistic image contents in the missing regions effectively.
Context Encoders~\cite{pathakCVPR16context} pioneers CNN-based image inpainting by proposing an adversarial network with an encoder-decoder architecture.
However, due to the information bottleneck layer of the autoencoder, the results are often blurry, incoherent, and can not work on irregular masks.
Yu~\etal~\cite{yu2018generative} proposed a coarse-to-fine network and a context-aware mechanism to reduce blurriness.
Iizuka~\etal~\cite{IizukaSIGGRAPH2017} adopted local and global discriminators and used dilated convolutions to increase the model's receptive field and enhance coherence.
Liu~\etal~\cite{liu2018partialinpainting} proposed partial convolutions, which only consider valid pixels during convolution, to handle irregular masks.
Yu~\etal~\cite{yu2018free} further extends the partial convolutions by introducing a dynamic feature gating mechanism, named gated convolutions, to deal with free-from masks.
Both Liu~\etal~\cite{liu2018partialinpainting} and Yu~\etal~\cite{yu2018free} adopt PatchGAN discriminator~\cite{pix2pix2017} to improve the coherence further.
%
Recently, several models were proposed to significantly improve the image painting quality by incorporating the structure knowledge in a different context, including image edges~\cite{nazeri2019edgeconnect,Li_2019_ICCV}, object contours~\cite{Xiong_2019_CVPR}, smooth edge-preserving map~\cite{ren2019structureflow}, and gradient map~\cite{jie2020inpainting}.
Nazeri~\etal~\cite{nazeri2019edgeconnect} introduced a two-stage network named EdgeConnect, which firstly recovers the missing edges in the masked regions, followed by a generator conditioned on the reconstructed edge map.
The authors prove that the structure-to-content approach can effectively preserve the structure information in the inpainting results. However, EdgeConnect uses canny edges to represent structure features, which might be suitable for natural images but may lead to complex local edges in indoor scenes. In contrast, our work exploits the HorizonNet~\cite{SunHSC19} to estimate layout edges, representing the global room structure, which is suitable for our indoor inpainting task. In addition, our model is an end-to-end architecture instead of a two-stage network.
Yang~\etal~\cite{jie2020inpainting} developed a multi-task learning framework to jointly learn the completion of image contents and structure map (edges and gradient).
A structure embedding scheme is employed to embed the learned structure features while inpainting explicitly.
The model further learns to exploit the recurrent structures and contents via an attention mechanism.
While demonstrating impressive performance in generating realistic results, these structure-aware methods still fail to model long-range structure correspondence such as the layout in the indoor scenes.
On the other hand, some works have successfully recovered a single partially occluded object~\cite{8578741,9710417}. However, their architecture does not handle multiple object instances of the same class and is thus not suitable for our context where the plane-wise segmentation consists of different numbers of wall planes.
\begin{figure*}[t]
    \begin{overpic}[width=\textwidth]{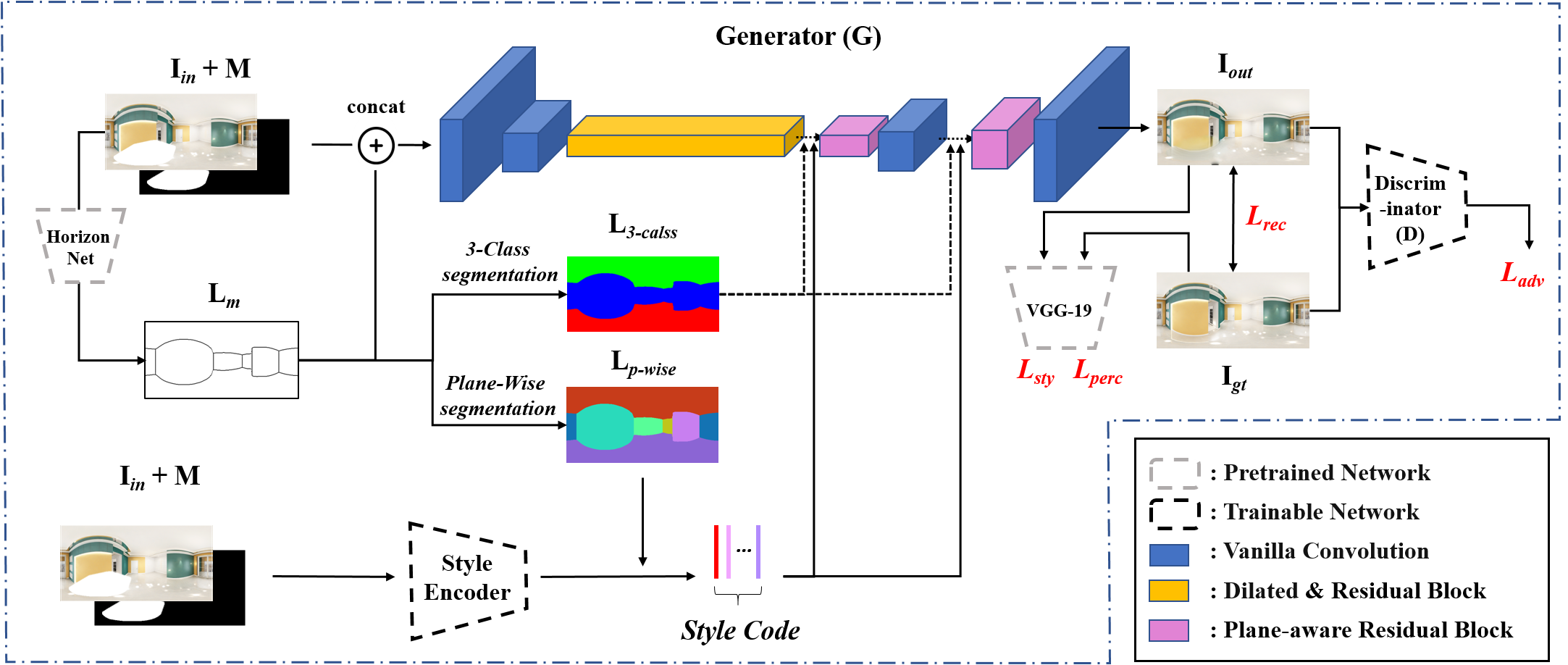}
    \end{overpic}
    \\
    \caption{\tb{Architecture overview.} Our network architecture follows the conventional generative adversarial network with an encoder-decoder scheme supervised by low- and high-level loss functions and a discriminator. Given a masked indoor panoramic image $\incompletedpano$ with a corresponding mask $\mask$, our system uses an off-the-shelf layout prediction network to predicts a layout map. The low-level boundary lines in $\layoutgmap$ serve as a conditional input to our network to assist the inpainting. Then, we compute two semantic segmentation maps from the layout map $\layoutgmap$, declared  $\layoutthreeclass$ and $\layoutplanewise$, where the latter is used to generate plane-wise style codes for ceiling, floor, and individual walls. Finally, these per plane style codes, together with $\layoutthreeclass$, are fed to a structural plane-aware normalization module to constrain the inpainting.
    }
\label{fig:overview}
\end{figure*}

\heading{Image-to-image translation.} 
The image inpainting is essentially a constrained image-to-image translation problem.
Significant efforts have been made to tackle various problems based on image-to-image translation architectures~\cite{pix2pix2017,CycleGAN2017,karras2019stylebased}.
Here we focus on the ones that are closely related to our work.
Park~\etal~\cite{park2019SPADE} introduced SPADE, which utilizes a spatial adaptive normalization layer for synthesizing photorealistic images given an input semantic segmentation map.
Specifically, a spatially-adaptive learned transform modulates the activation layer with a semantic segmentation map and effectively propagates the semantic information throughout the network.
In contrast to SPADE, which uses only one style code to control the image synthesis, Zhu~\etal~\cite{Zhu_2020_CVPR} presents SEAN by extending the SPADE architecture with per-region style encoding.
By embedding one style code for individual semantic classes, SEAN shows significant improvement over SPADE and generates the highest quality results. 
In the context of indoor scenes, Gkitsas~\etal~\cite{gkitsas2021panodr} introduce PanoDR that combines image-to-image translation with a generator to constrain the image inpainting with the underlying scene structure.
Percisely, to convert a furnished indoor panorama into its empty counterpart, PanoDR exploits a generator for synthesizing photorealistic image contents where the global layout structure is preserved via an image-to-image translation module.
The empty indoor panorama is then used to complete the masked regions in the input panorama via a simple copy-and-paste process.
Gkitsas~\etal~\cite{DBLP:conf/visapp/GkitsasZSDZ22} extend the architecture of PanoDR to make the model end-to-end trainable. However, the quantitative evaluation indicates that the performance improvement is marginal compared with PanoDR.
Our system also combines a generator with image-to-image translation as PanoDR does. However, we obtain superior results than PanoDR by exploiting the global layout edges as a prior and adapting SEAN blocks in a local plane-wise manner to guide the inpainting. Moreover, in contrast to PanoDR performs the
inpainting task via an indirect way, our system performs the inpainting task in an end-to-end fashion, directly completing the mask areas instead of hallucinating an empty scene, thus resulting in better visual quality and consistency.
\section{Overview}
\label{sec:overview}

\figref{overview} illustrates an overview of our architecture.
Our system takes a masked panoramic image $\incompletedpano$ and the corresponding binary mask $\mask$ as inputs and generates the inpainted panoramic image $\inpaintedpano$.
The masked panoramic image is generated by ${\incompletedpano = \groundtruthpano \odot(\mathbf{1}-\mask)}$, where $\groundtruthpano$ represents the ground-truth panoramic image and $\odot$ denotes the Hadamard product.
Our system first utilizes an off-the-shelf model to estimate the room layout $\layoutgmap$ from input masked panoramic image.
This layout map is then concatenated with $\incompletedpano$ and $\mask$ to obtain a five-channel input map fed into the generator $\generator$.
We further derive two semantic segmentation maps $\layoutthreeclass$ and $\layoutplanewise$ using the layout map for the subsequent normalization module (\secref{layout_guidance}).
The image generation model follows the conventional generative adversarial architecture with one content encoder and one image decoder with one discriminator. (\secref{inpainting_backbone}).
%
To impose structure information during inpainting, we introduce a plane-aware normalization that modifies the SEAN~\cite{Zhu_2020_CVPR} block with two semantic segmentation maps to guide the decoder with spatially varying normalization parameters per structural plane (\ie ceiling, floor, and individual walls).
Such a plane-aware normalization provides useful guidance for global structure preservation as well as consistent local image content generation (\secref{planar-aware_module}).
Finally, common loss functions in image inpainting, including the reconstruction loss, the perceptual loss, the style loss, and the adversarial loss are employed to train our model (\secref{loss_func}).
\section{Method}
\label{sec:method}
\begin{figure}[!t]
    \centering
    \begin{overpic}[width=\columnwidth]{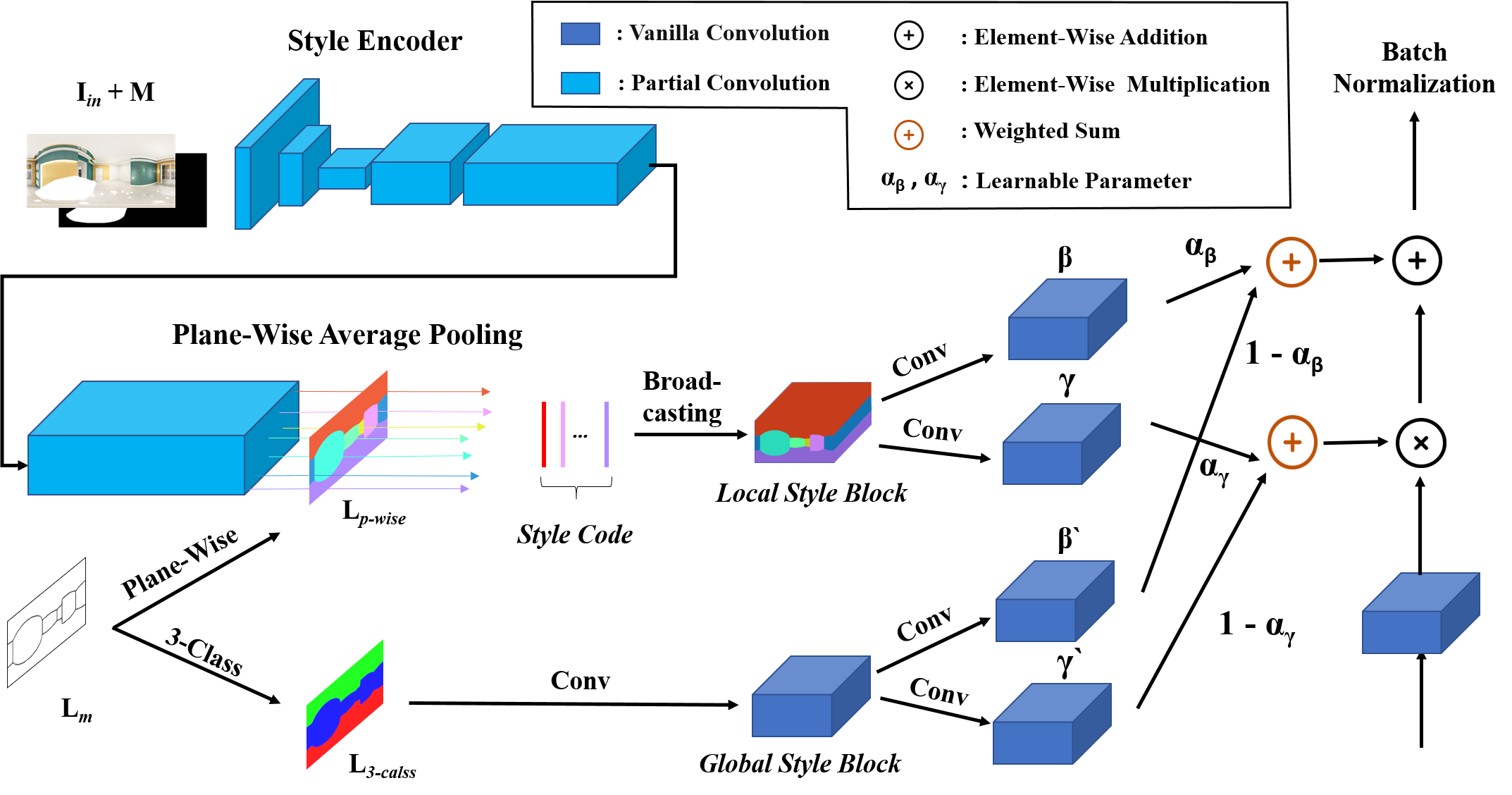}
    \end{overpic}
    \\
    \caption{\tb{Plane-aware normalization.}  Given an incomplete indoor panoramic image $\incompletedpano$ with mask $\mask$, we first predict two normalization values $\beta$ and $\gamma$ through several partial convolution~\cite{liu2018partialinpainting} blocks and a plane-wise average pooling based on the plane-wise segmentation map $\layoutplanewise$. Second, we predict another set of normalization values $\beta'$ and $\gamma'$ through several vanilla convolution blocks based on the 3-class segmentation map $\layoutthreeclass$. The final normalization values are thus computed using the weighted sum weighted by learnable parameters $\alpha_{\beta}$ and $\alpha_{\gamma}$.
    }
\label{fig:pa_normal}
\end{figure}

\subsection{Layout Guidance Map}
\label{sec:layout_guidance}
We employ an off-the-shelf model, HorizonNet~\cite{SunHSC19}, to estimate a layout map from input masked panorama.
Through a recurrent neural network, the HorizonNet predicts a 3-dimensional vector representing ceiling, ground, and corner location.
We further process the output vector to generate a layout map $\layoutgmap$ comprising low-level boundary lines.
This layout map serves as a conditional input to encourage the preservation of global layout structure while inpainting.
Moreover, we extract two semantic segmentation maps from the layout map that depict
(i) the segmentation mask $\layoutthreeclass$ with three semantic labels of indoor scene, \ie ceiling, floor, and wall; and
(ii) a plane-wise segmentation mask $\layoutplanewise$ where pixels are indexed in a per structural plane basis (\ie ceiling, floor, or individual walls).
These semantic segmentation maps are generated using conventional image processing operations (i.e, flood-fill) and will be used in the later normalization module.

\subsection{Image Inpainting Backbone}
\label{sec:inpainting_backbone}
As shown in~\figref{overview}, our network architecture consists of one generator and one discriminator. 
The generator $\generator$ follows a conventional scheme with one content encoder and one image decoder.
The content encoder consists of two down-sampling convolution blocks followed by eight residual blocks using dilated convolution~\cite{7780459}.
The image decoder uses a cascade of our proposed plane-aware residual blocks and two up-sampling blocks.
Motivated by EdgeConnect~\cite{nazeri2019edgeconnect}, we use PatchGAN~\cite{isola2018imagetoimage} as our discriminator to determine the real or fake sample by dividing the input image into several patches.
In the following sections, we will elaborate plane-aware residual block, loss functions, and discriminator in more detail.

\subsection{Plane-aware Normalization}
\label{sec:planar-aware_module}
Considering the different styles among wall planes is very common in real-world indoor scenes.
We follow the architecture of SEAN~\cite{Zhu_2020_CVPR} and propose leveraging two kinds of segmentation maps $\layoutplanewise$ and $\layoutthreeclass$ to establish our plane-aware normalization (see~\figref{pa_normal}).
Our plane-aware normalization consists of one style encoder and two style blocks, which enhance the global style semantics and local style consistency of the generated results.
The inputs of the style encoder include masked panoramic image $\incompletedpano$ and mask image $\mask$. 
We use partial convolution blocks in style encoder instead of vanilla convolution to make feature extraction conditioned only valid pixels.
We first adopt the plane-wise average pooling on the output features to generate style codes for each plane based on $\layoutplanewise$.
Second, we spatially broadcast each style code on the corresponding area and output the local style block.
On the other side, we predict the global style block by passing the 3-class segmentation map $\layoutthreeclass$ through several convolution layers.
Finally, the remaining part of our plane-aware normalization follows the same architecture of SEAN~\cite{Zhu_2020_CVPR}, and combines global and local style blocks into the downstream $\beta$ and $\gamma$ parameters of the final batch normalization.

\subsection{Loss Functions}
\label{sec:loss_func}
Here we elaborate on the low- and high-level loss functions and the discrimination used for training our image generator.

\heading{Reconstruction loss} measures the low-level pixel-based loss between the predicted and ground-truth images.
To encourage the generator to pay more attention to the missing regions, we additionally calculate the $\oneloss$ loss in the missing regions. The reconstruction loss $\recloss$ is defined as follows:

\begin{equation}
\recloss=\left\|\mask\odot \groundtruthpano - \mask\odot \inpaintedpano \right\|_1+\left\|\groundtruthpano - \inpaintedpano \right\|_1,
\label{eqn:rec_loss}
\end{equation}
where $\groundtruthpano$ and $\inpaintedpano$ represent the ground-truth image and the generator's output, respectively, and $\mask$ is a binary mask.

\heading{Perceptual loss} encourages the predicted and ground-truth images to have similar representation in high-level feature space extracted via a pre-trained VGG-19~\cite{simonyan2015deep}, and is defined as follows:
\begin{equation}
\percloss=\sum_{i} \left\|\phi_{i}\left(\groundtruthpano\right)-\phi_{i}\left(\inpaintedpano\right)\right\|_{1},
\label{eqn:perc_loss}
\end{equation}
where $\phi_{i}$ is the activation map of the $ith$ layer of the pre-trained feature extraction network. 

\heading{Style loss} calculates the co-variance difference between the activation maps. For the activation map $\phi_{i}$ of size $C_{i}$ × $H_{i}$ × $W_{i}$, the style loss is defined as follows:
\begin{equation}
\styloss=\left\|G_{i}^{\phi}\left(\groundtruthpano\right)-G_{i}^{\phi}\left(\inpaintedpano\right)\right\|_{1},
\label{eqn:sty_loss}
\end{equation}
where $G_{i}^{\phi}$ is a $C_{i}$ × $C_{i}$ gram matrix~\cite{Gatys_2016_CVPR} constructed by the activation map $\phi_{i}$.

\heading{Adversarial loss} is implemented with the patch-based discriminator~\cite{isola2018imagetoimage}, which outputs the feature map divided into several feature patches and uses hinge loss~\cite{lim2017geometric} to optimize the generator $G$ and the discriminator $D$.
The adversarial loss for generator $G$ and discriminator $D$ are defined as follows:
\begin{equation}
L_G=-D\left(\inpaintedpano \right),
\label{eqn:adv_loss}
\end{equation}
\begin{equation}
L_D=\lambda_{D}\left(max\left(0,1+D \left(\inpaintedpano \right)\right) + max\left(0,1-D \left(\groundtruthpano \right)\right)\right);
\label{eqn:d_loss}
\end{equation}
The overall loss function used in the generator $G$ is defined as follows: 
\begin{equation}
\totalloss=\lambda_{rec} \recloss+\lambda_{perc} \percloss+\lambda_{sty} \styloss+\lambda_{G} L_G,
\label{eqn:total_loss}
\end{equation}
where $\lambda_{rec}$, $\lambda_{perc}$, $\lambda_{sty}$, $\lambda_{G}$, and $\lambda_{D}$ are the hyperparameters for weighting the loss functions.

\begin{figure*}[]
\centering
    \begin{subfigure}[t]{.16\textwidth}
        \includegraphics[width=\textwidth,keepaspectratio]{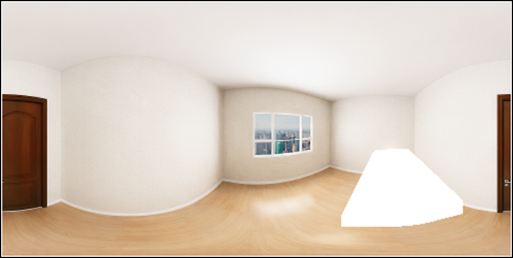} 
        \includegraphics[width=\textwidth,keepaspectratio]{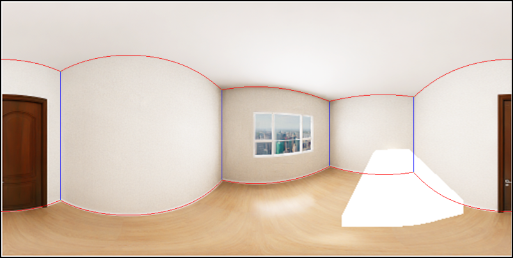} 
       \includegraphics[width=\textwidth,keepaspectratio]{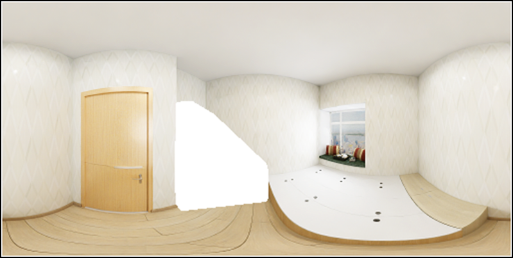} 
       \includegraphics[width=\textwidth,keepaspectratio]{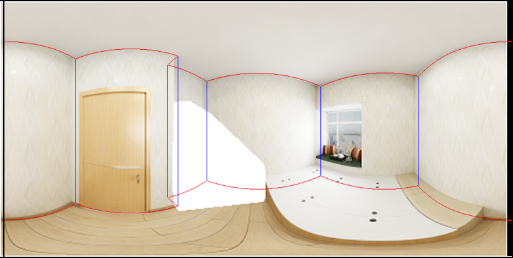} 
       \includegraphics[width=\textwidth,keepaspectratio]{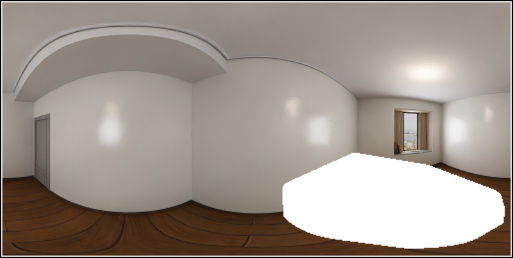} 
       \includegraphics[width=\textwidth,keepaspectratio]{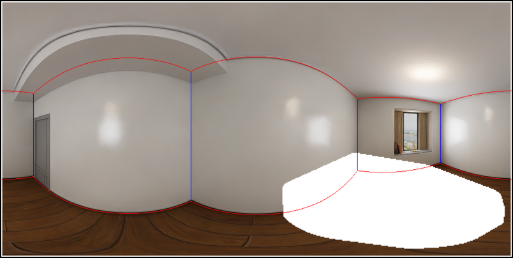} 
       \includegraphics[width=\textwidth,keepaspectratio]{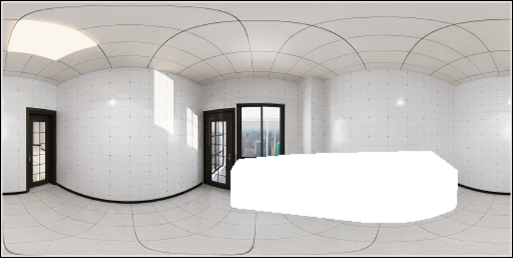} 
       \includegraphics[width=\textwidth,keepaspectratio]{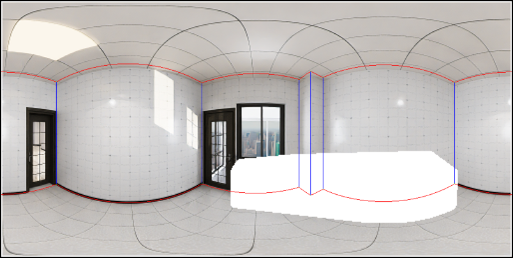} 
        \includegraphics[width=\textwidth,keepaspectratio]{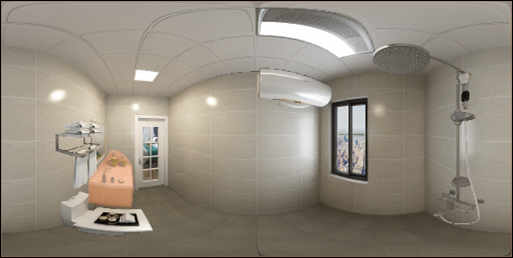} 
        \includegraphics[width=\textwidth,keepaspectratio]{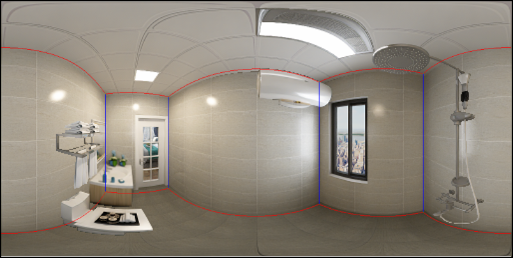} 
       \includegraphics[width=\textwidth,keepaspectratio]{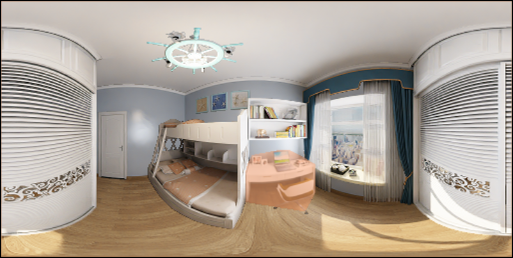} 
       \includegraphics[width=\textwidth,keepaspectratio]{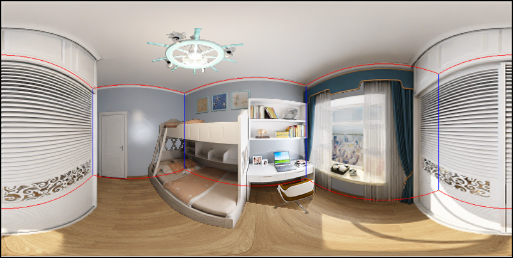} 
       \includegraphics[width=\textwidth,keepaspectratio]{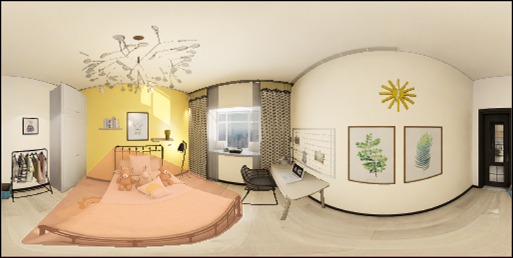} 
       \includegraphics[width=\textwidth,keepaspectratio]{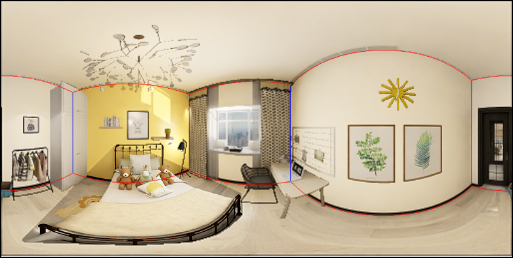} 
       \includegraphics[width=\textwidth,keepaspectratio]{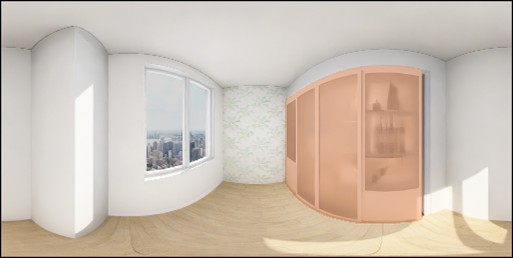} 
       \includegraphics[width=\textwidth,keepaspectratio]{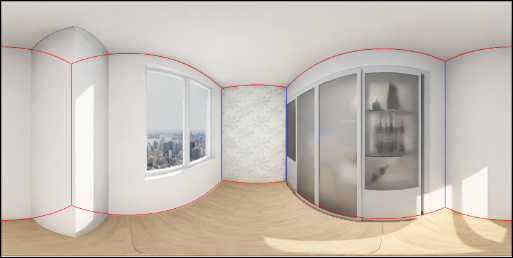} 

       \caption*{Input / Layout}
    \end{subfigure}
    \begin{subfigure}[t]{.16\textwidth}
        \includegraphics[width=\textwidth,keepaspectratio]{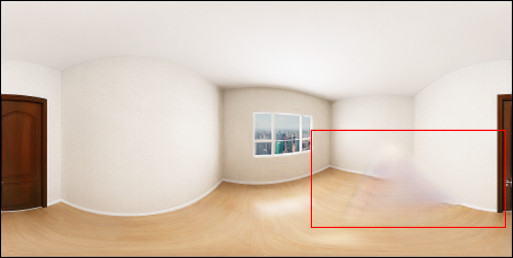} 
        \includegraphics[width=\textwidth,keepaspectratio]{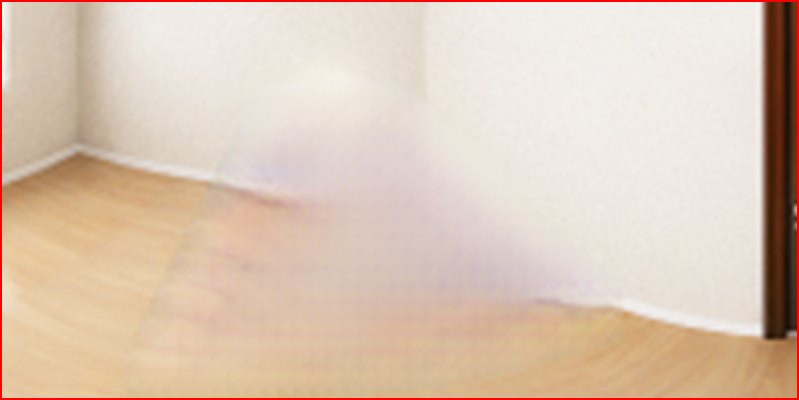} 
       \includegraphics[width=\textwidth,keepaspectratio]{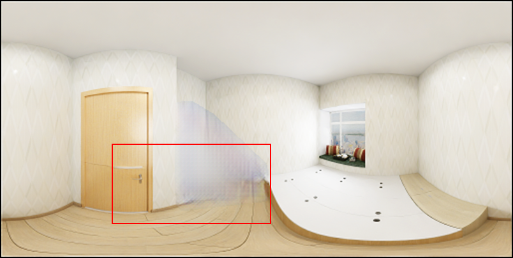} 
       \includegraphics[width=\textwidth,keepaspectratio]{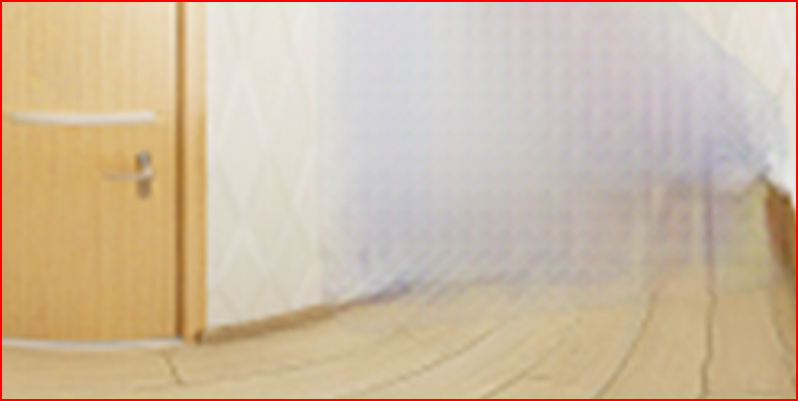} 
       \includegraphics[width=\textwidth,keepaspectratio]{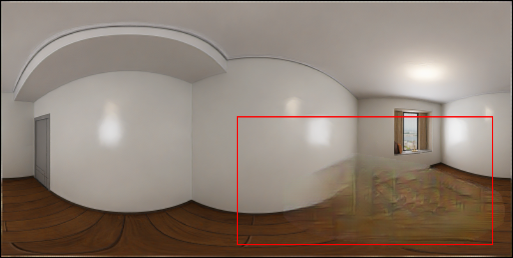} 
       \includegraphics[width=\textwidth,keepaspectratio]{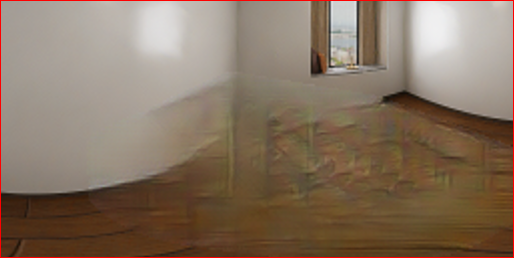} 
       \includegraphics[width=\textwidth,keepaspectratio]{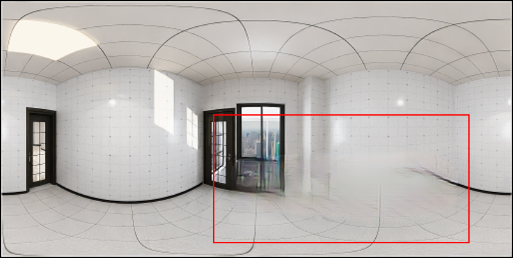} 
       \includegraphics[width=\textwidth,keepaspectratio]{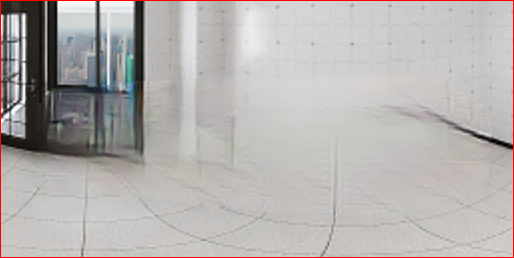} 
        \includegraphics[width=\textwidth,keepaspectratio]{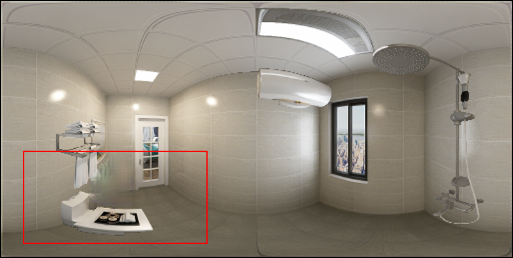} 
        \includegraphics[width=\textwidth,keepaspectratio]{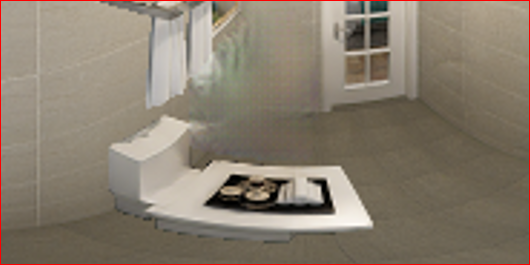} 
       \includegraphics[width=\textwidth,keepaspectratio]{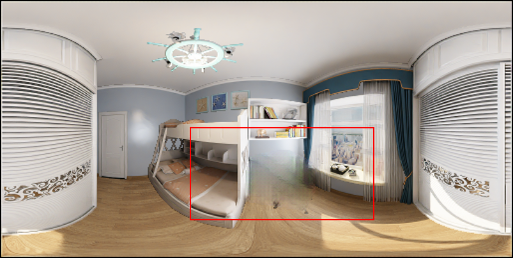} 
       \includegraphics[width=\textwidth,keepaspectratio]{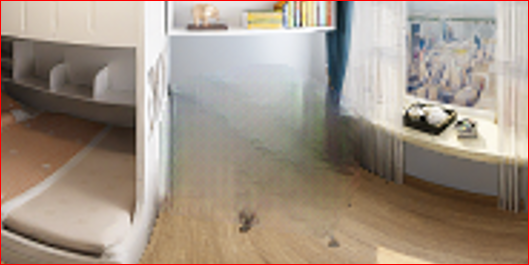} 
       \includegraphics[width=\textwidth,keepaspectratio]{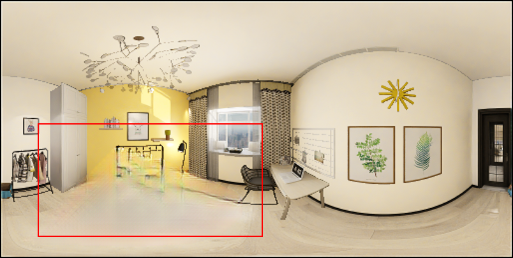} 
       \includegraphics[width=\textwidth,keepaspectratio]{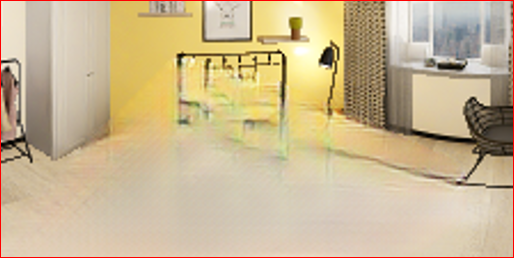} 
       \includegraphics[width=\textwidth,keepaspectratio]{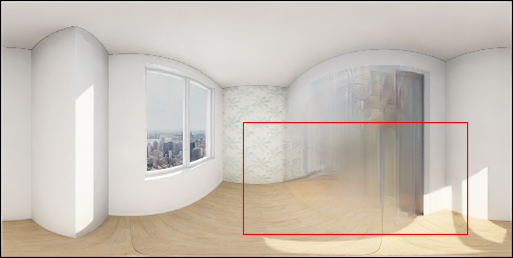} 
       \includegraphics[width=\textwidth,keepaspectratio]{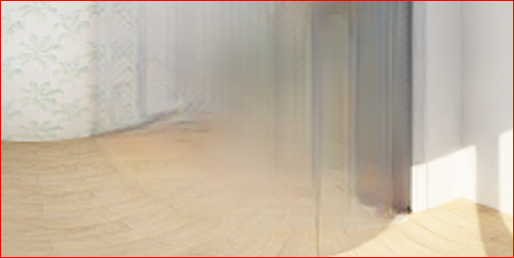} 

        \caption*{EC~\cite{nazeri2019edgeconnect}}
       
    \end{subfigure}
    \begin{subfigure}[t]{.16\textwidth}
        \includegraphics[width=\textwidth,keepaspectratio]{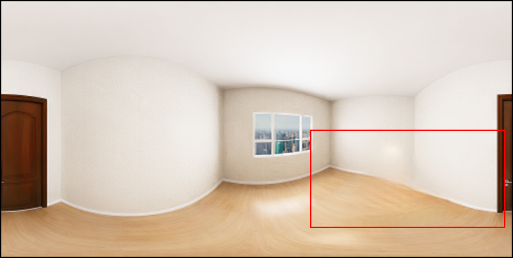} 
        \includegraphics[width=\textwidth,keepaspectratio]{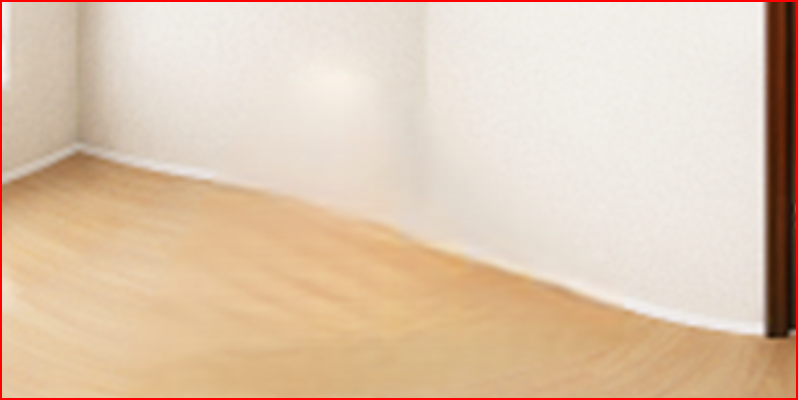} 
       \includegraphics[width=\textwidth,keepaspectratio]{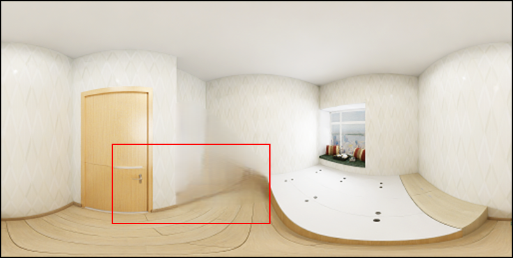} 
       \includegraphics[width=\textwidth,keepaspectratio]{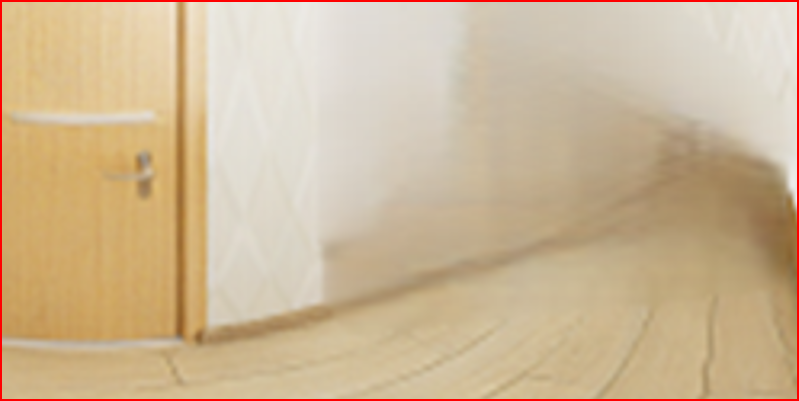} 
       \includegraphics[width=\textwidth,keepaspectratio]{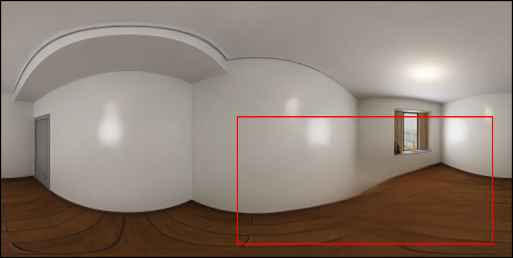} 
       \includegraphics[width=\textwidth,keepaspectratio]{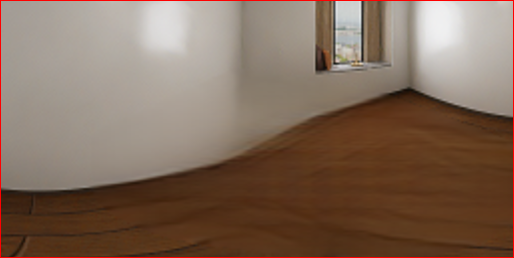} 
       \includegraphics[width=\textwidth,keepaspectratio]{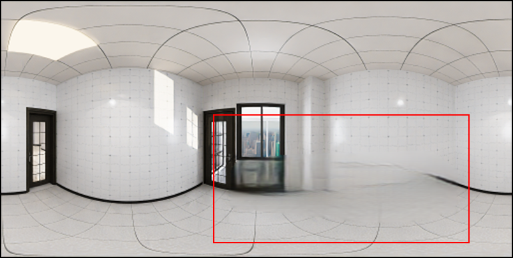} 
       \includegraphics[width=\textwidth,keepaspectratio]{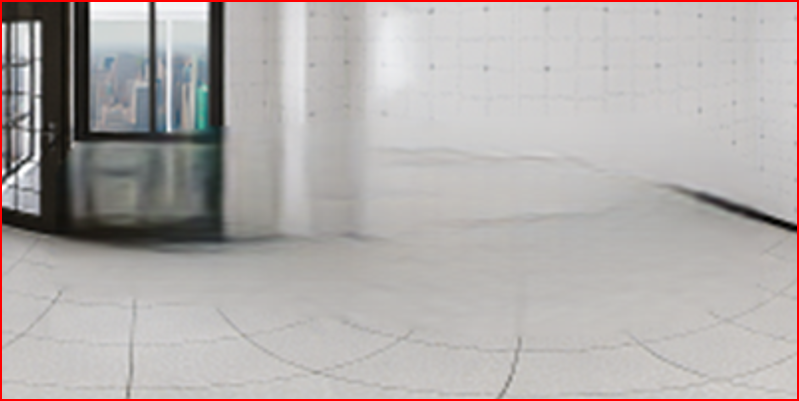} 
        \includegraphics[width=\textwidth,keepaspectratio]{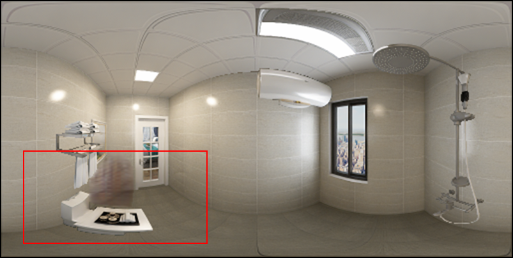} 
        \includegraphics[width=\textwidth,keepaspectratio]{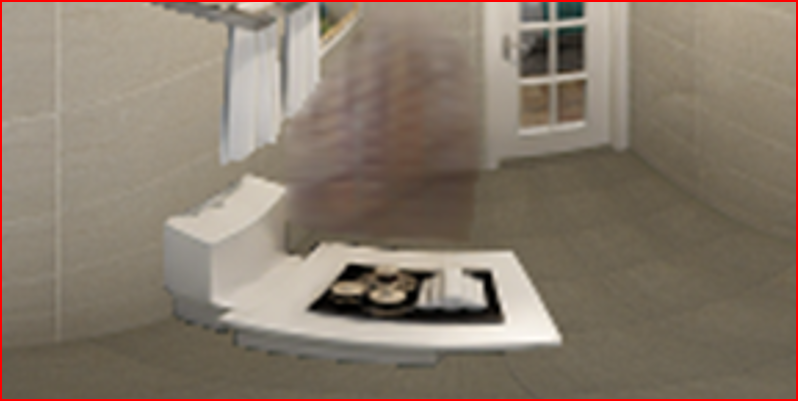} 
       \includegraphics[width=\textwidth,keepaspectratio]{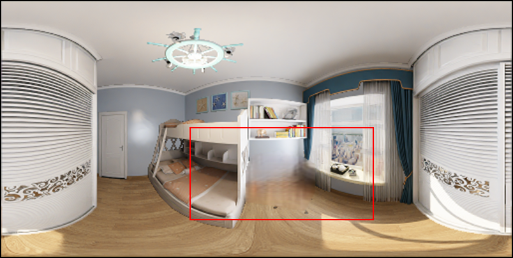} 
       \includegraphics[width=\textwidth,keepaspectratio]{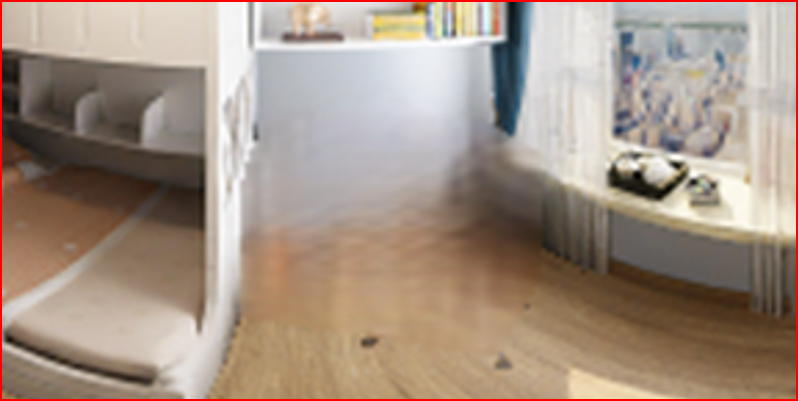} 
       \includegraphics[width=\textwidth,keepaspectratio]{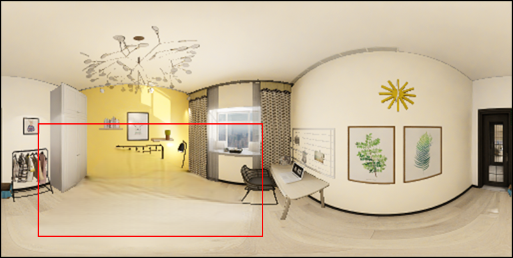} 
       \includegraphics[width=\textwidth,keepaspectratio]{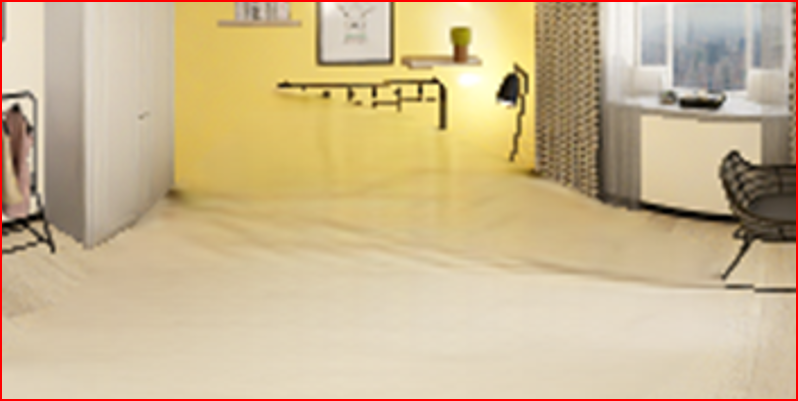} 
       \includegraphics[width=\textwidth,keepaspectratio]{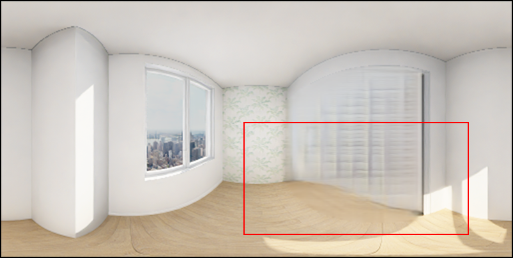} 
       \includegraphics[width=\textwidth,keepaspectratio]{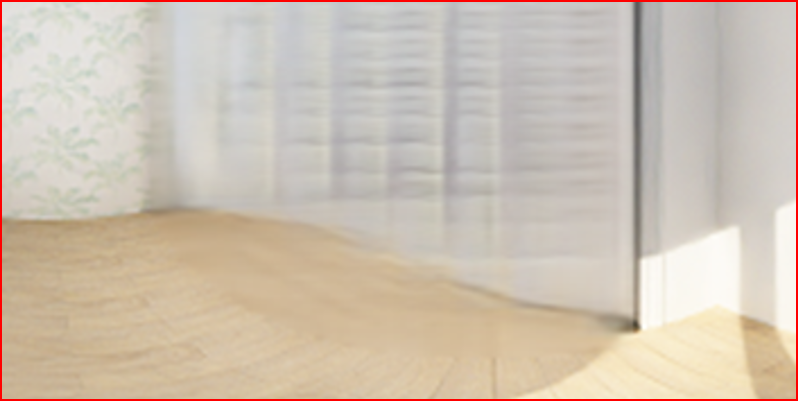} 

       \caption*{LISK~\cite{jie2020inpainting}}
    \end{subfigure}
    \begin{subfigure}[t]{.16\textwidth}
        \includegraphics[width=\textwidth,keepaspectratio]{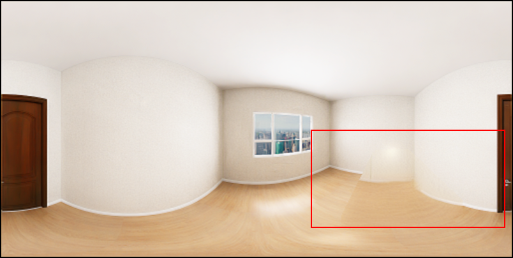} 
        \includegraphics[width=\textwidth,keepaspectratio]{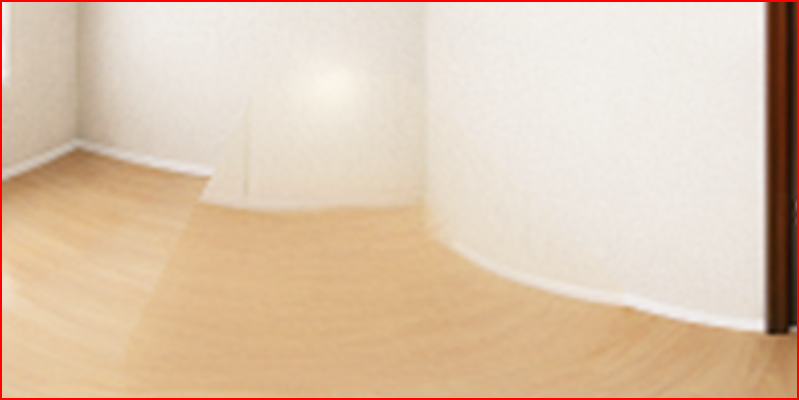} 
       \includegraphics[width=\textwidth,keepaspectratio]{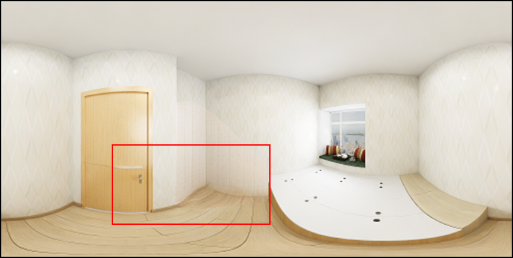} 
       \includegraphics[width=\textwidth,keepaspectratio]{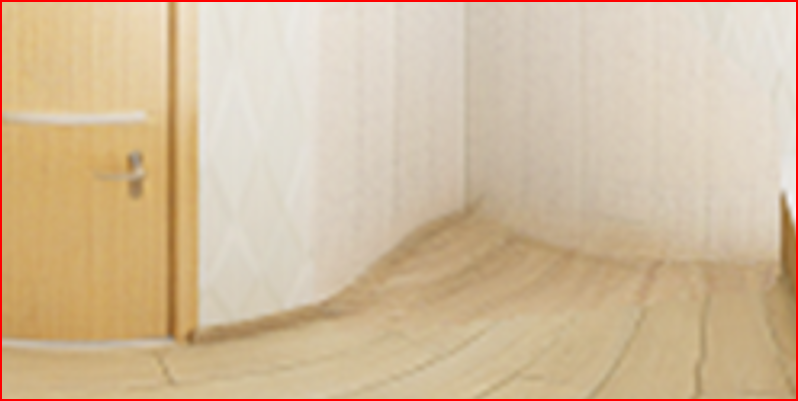} 
       \includegraphics[width=\textwidth,keepaspectratio]{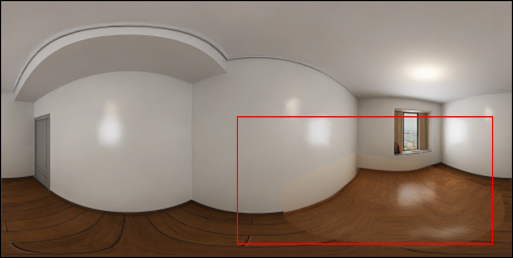} 
       \includegraphics[width=\textwidth,keepaspectratio]{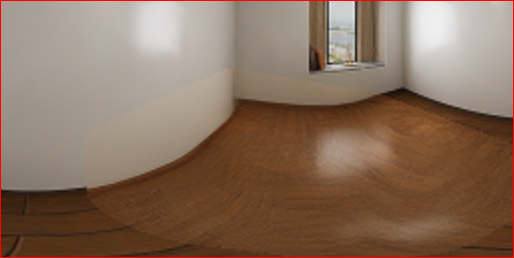} 
       \includegraphics[width=\textwidth,keepaspectratio]{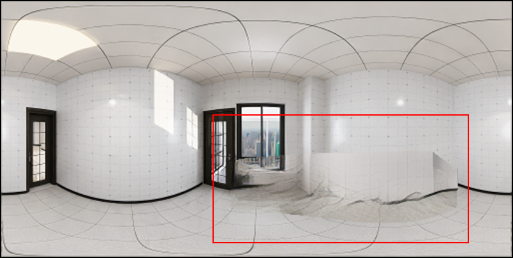} 
       \includegraphics[width=\textwidth,keepaspectratio]{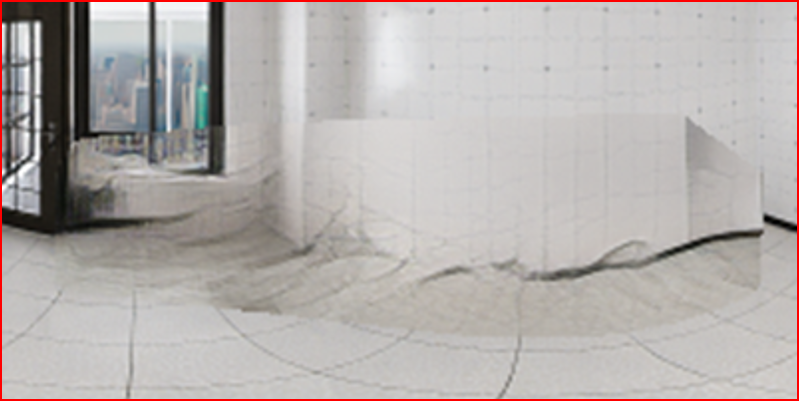} 
        \includegraphics[width=\textwidth,keepaspectratio]{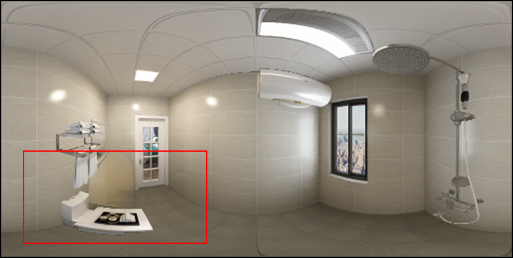} 
        \includegraphics[width=\textwidth,keepaspectratio]{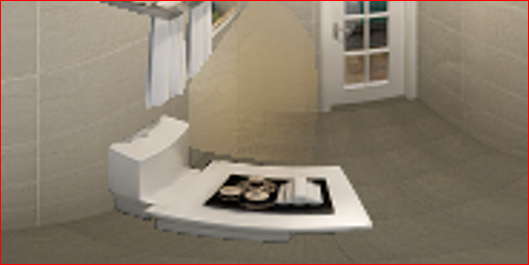} 
       \includegraphics[width=\textwidth,keepaspectratio]{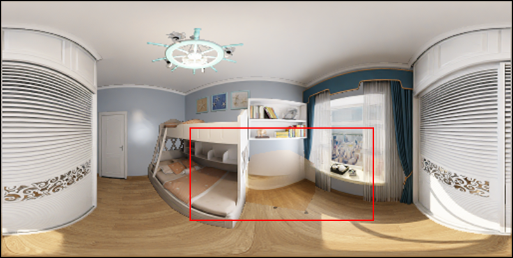} 
       \includegraphics[width=\textwidth,keepaspectratio]{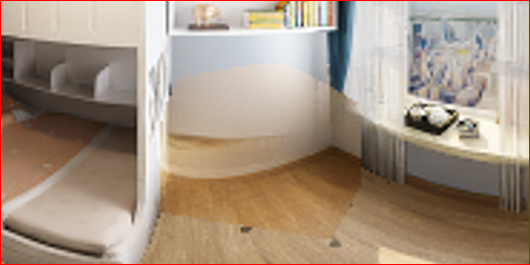} 
       \includegraphics[width=\textwidth,keepaspectratio]{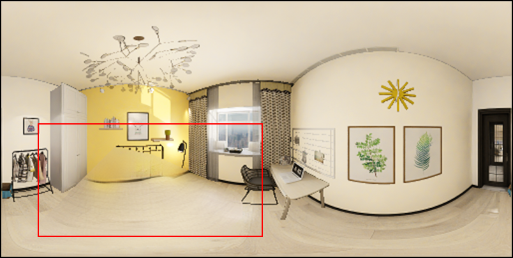} 
       \includegraphics[width=\textwidth,keepaspectratio]{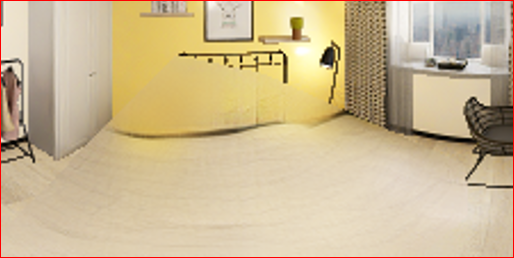} 
       \includegraphics[width=\textwidth,keepaspectratio]{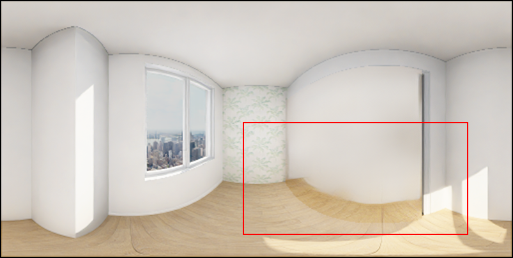} 
       \includegraphics[width=\textwidth,keepaspectratio]{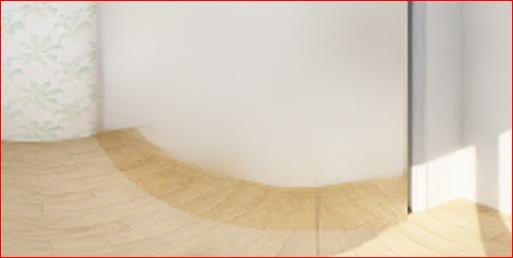} 

       \caption*{PanoDR~\cite{gkitsas2021panodr}}
    \end{subfigure}
    \begin{subfigure}[t]{.16\textwidth}
        \includegraphics[width=\textwidth,keepaspectratio]{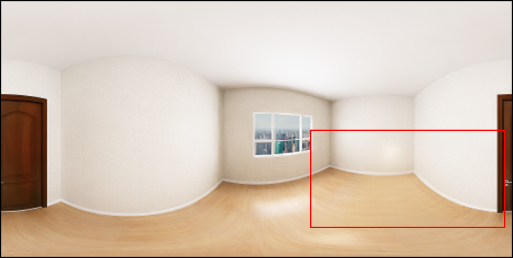} 
        \includegraphics[width=\textwidth,keepaspectratio]{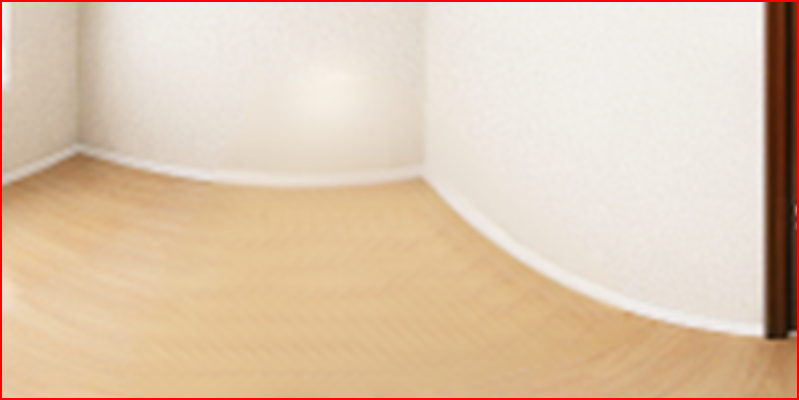} 
       \includegraphics[width=\textwidth,keepaspectratio]{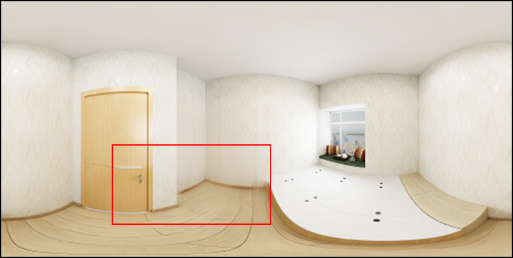} 
       \includegraphics[width=\textwidth,keepaspectratio]{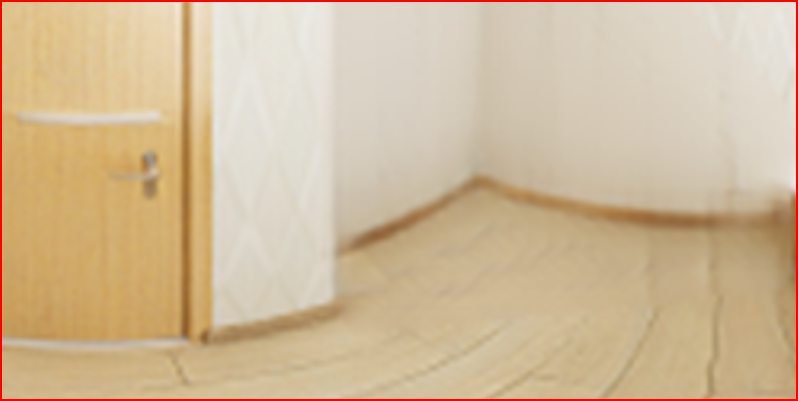} 
       \includegraphics[width=\textwidth,keepaspectratio]{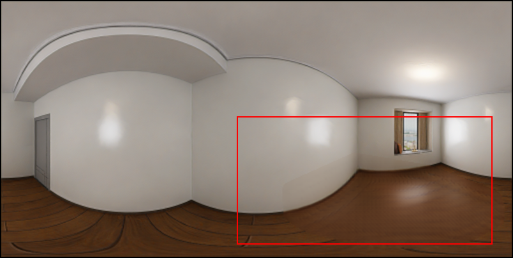} 
       \includegraphics[width=\textwidth,keepaspectratio]{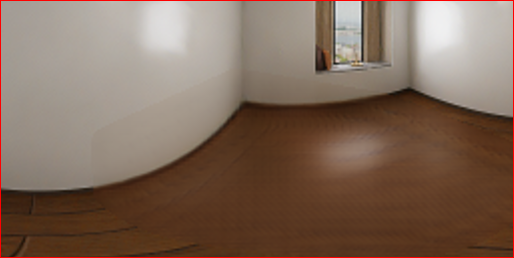} 
       \includegraphics[width=\textwidth,keepaspectratio]{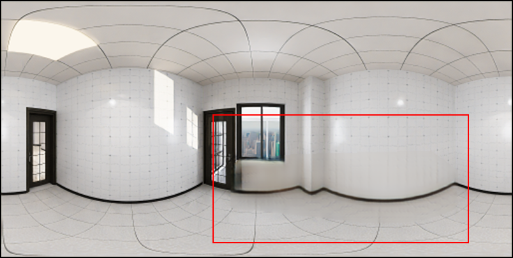} 
       \includegraphics[width=\textwidth,keepaspectratio]{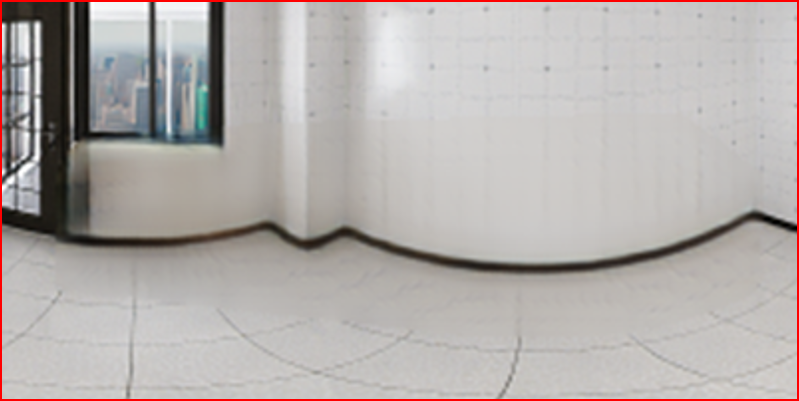} 
        \includegraphics[width=\textwidth,keepaspectratio]{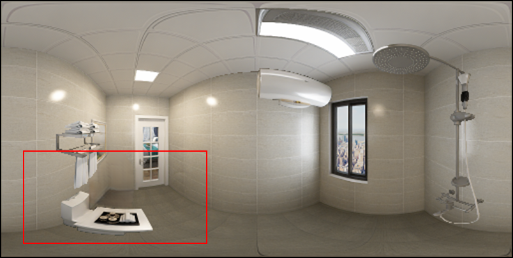} 
        \includegraphics[width=\textwidth,keepaspectratio]{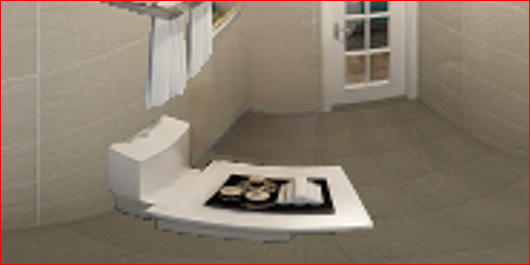} 
       \includegraphics[width=\textwidth,keepaspectratio]{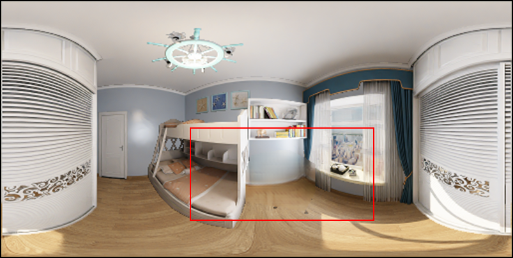} 
       \includegraphics[width=\textwidth,keepaspectratio]{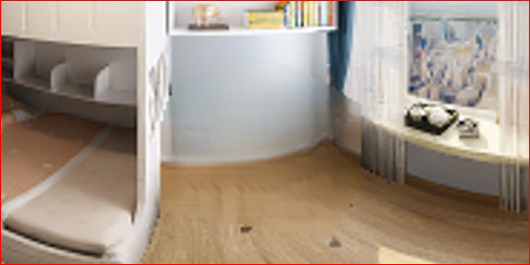} 
       \includegraphics[width=\textwidth,keepaspectratio]{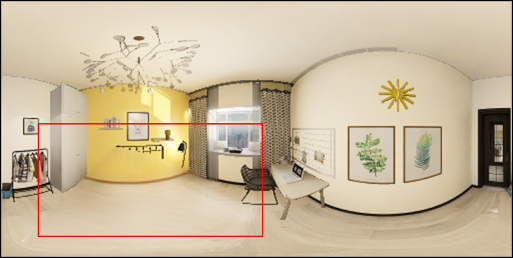} 
       \includegraphics[width=\textwidth,keepaspectratio]{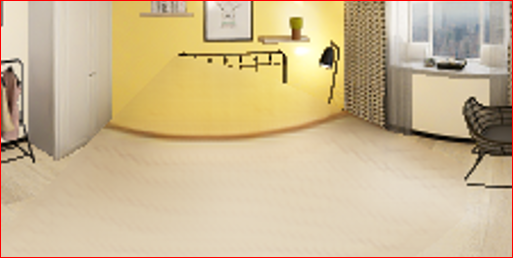} 
       \includegraphics[width=\textwidth,keepaspectratio]{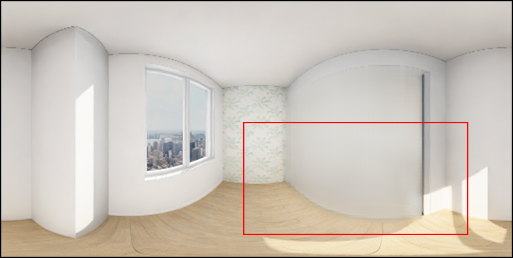} 
       \includegraphics[width=\textwidth,keepaspectratio]{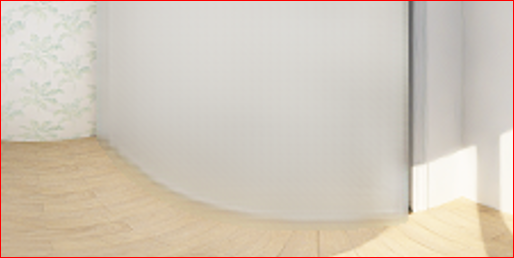} 
       
       \caption*{Ours}
    \end{subfigure}
    \begin{subfigure}[t]{.16\textwidth}
        \includegraphics[width=\textwidth,keepaspectratio]{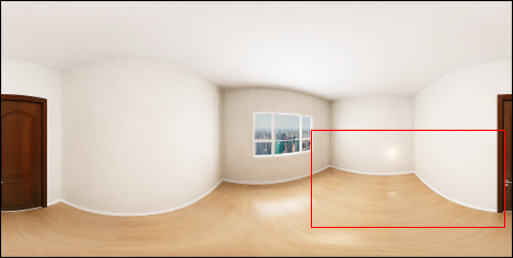} 
        \includegraphics[width=\textwidth,keepaspectratio]{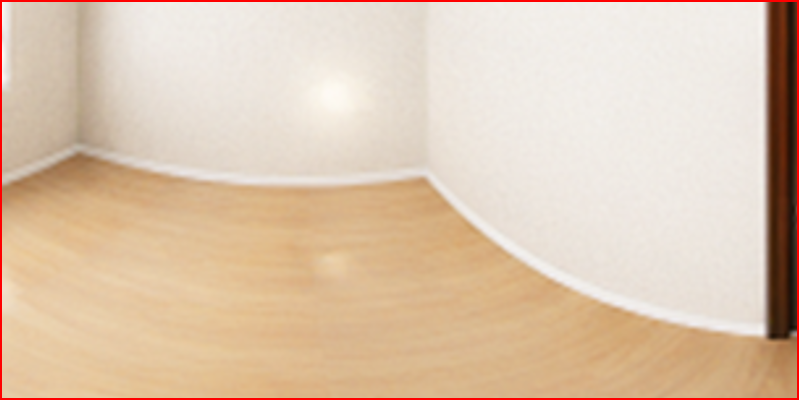} 
       \includegraphics[width=\textwidth,keepaspectratio]{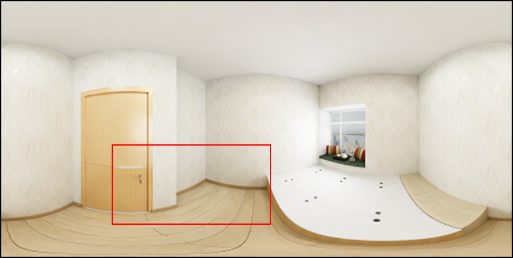} 
       \includegraphics[width=\textwidth,keepaspectratio]{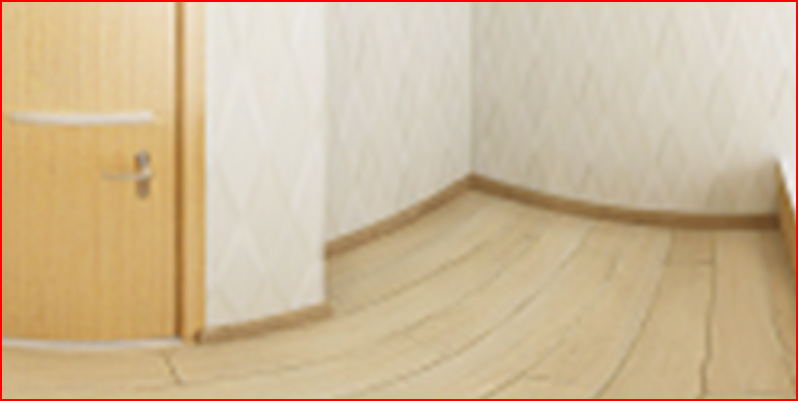} 
       \includegraphics[width=\textwidth,keepaspectratio]{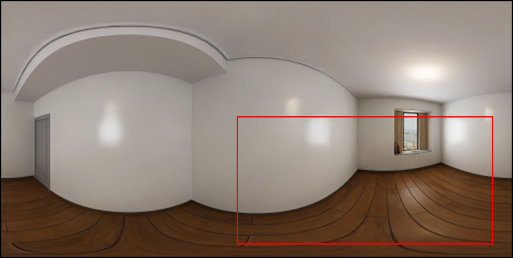} 
       \includegraphics[width=\textwidth,keepaspectratio]{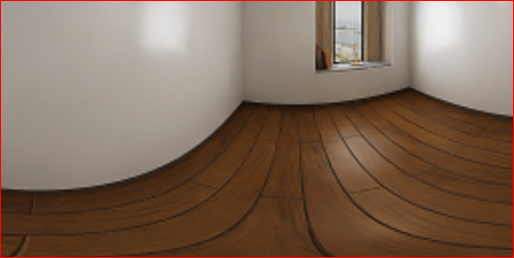} 
       \includegraphics[width=\textwidth,keepaspectratio]{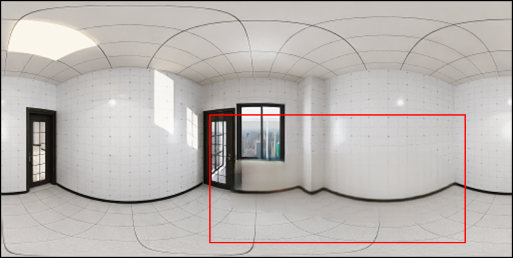} 
       \includegraphics[width=\textwidth,keepaspectratio]{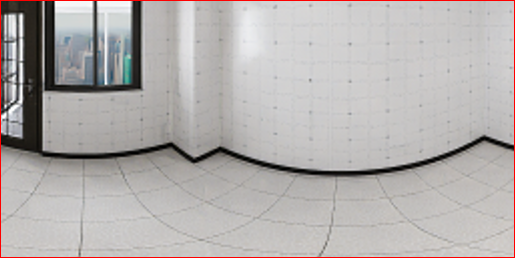} 
      \includegraphics[width=\textwidth,keepaspectratio]{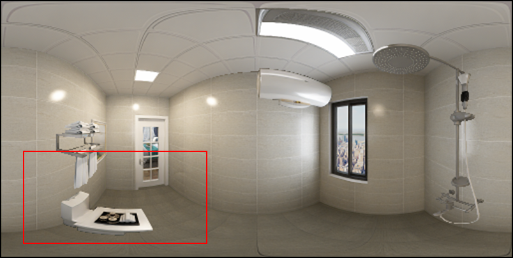} 
      \includegraphics[width=\textwidth,keepaspectratio]{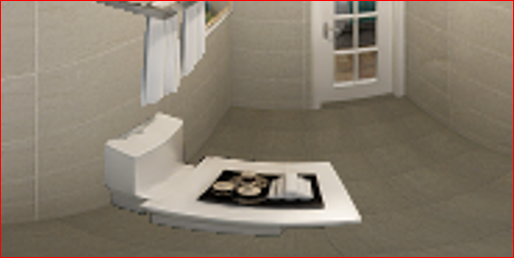} 
       \includegraphics[width=\textwidth,keepaspectratio]{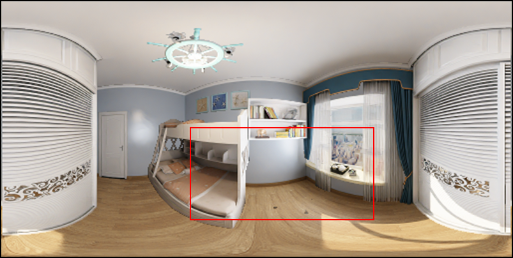} 
       \includegraphics[width=\textwidth,keepaspectratio]{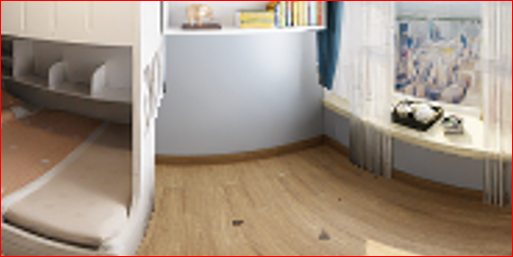} 
       \includegraphics[width=\textwidth,keepaspectratio]{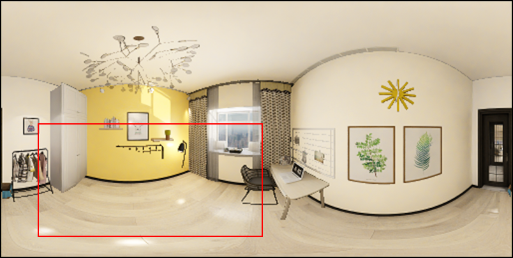} 
       \includegraphics[width=\textwidth,keepaspectratio]{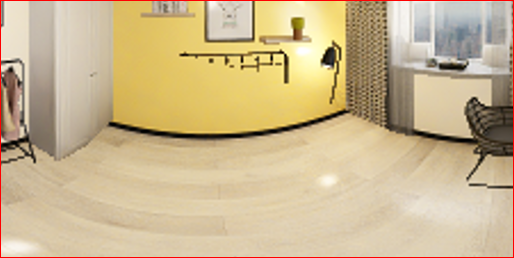} 
       \includegraphics[width=\textwidth,keepaspectratio]{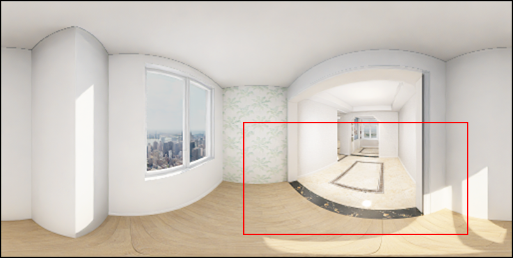} 
       \includegraphics[width=\textwidth,keepaspectratio]{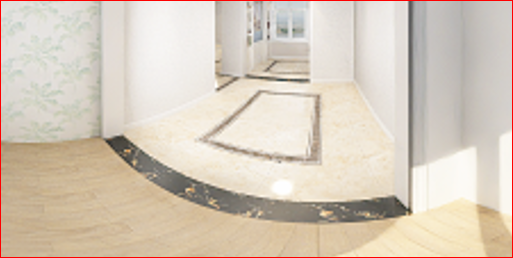} 
       
       \caption*{GT}
    \end{subfigure}
    \vspace{\figcapmargin}
    \caption{\textbf{Qualitative comparisons with state-of-the-arts.} 
    Top 8 rows: the inpainting results of the empty indoor scenes. Bottom 8 rows: the inpainting results of the furnished indoor scenes. Our method produces superior results in generating image contents that align the layout structure well and are consistent with the surrounding of the masked regions. 
    }
    \label{fig:comp_sota_si_fr}
\end{figure*}

\section{Experiments}
\label{sec:results}
In this section,  we evaluate the performance of our model by comparing it with several state-of-the-art image inpainting approaches and conducting ablation studies to verify the necessity of individual components in the proposed architecture.
Please refer to our online webpage for other experiments and more results\footnote{\url{https://ericsujw.github.io/LGPN-net/}}.

\subsection{Experimental Settings}
\heading{Dataset and baselines.}
We compare our model with the following state-of-the-art structure-aware image inpainting models:
\begin{itemize}
    \item EC~\cite{nazeri2019edgeconnect}: a two-stage adversarial network that comprises an edge completion model followed by a generator.
    
    \item LISK~\cite{jie2020inpainting}: a multi-task learning framework that exploits image structure embedding and an attention mechanism in the generator.
    
    \item PanoDR~\cite{gkitsas2021panodr}: a deep learning framework that combines image-to-image translation with generator to condition the inpainting on the indoor scene structure.
\end{itemize}
The experiments were conducted on a public indoor panorama dataset, Structured3D~\cite{Structured3D}, which contains 21,835 indoor panoramas.
The official data split is adopted for training(18,362), validation(1,776), and testing(1,697).
We follow the same procedure as PanoDR to generate mask images using contours of foreground furniture (see Section 3.1).
We use the officially released implementation of baselines for training from scratch and testing.
Note that each indoor panorama in Structured3D has two representations of the same scene (\ie empty and furnished).
Therefore, the experiments were conducted in two phases to evaluate our model and baselines in different application scenarios (\ie structural inpainting vs. furniture removal). 

\heading{Evaluation metrics.}
We take several commonly used image quality assessment metrics in previous inpainting approaches for quantitative evaluation.
Specifically, we used the low-level feature metrics, including Mean Absolute Error (MAE), Peak Signal-to-Noise (PSNR), Structural Similarity Index (SSIM)~\cite{1284395}, 
and Fréchet Inception Distance (FID)~\cite{heusel2018gans}. 

\heading{Implementation details.}
We implement our model in PyTorch and conduct the experiments on a single NVIDIA V100 with 32G VRAM. 
The resolution of the panoramic images is resized to $512 \times 256$.
We use Adam~\cite{kingma2017adam} optimizer in the training process with the hyper-parameters setting of $b_{1} = 0.0$ and $b_{2} = 0.9$, a learning rate of $0.0001$, and a batch size of 8.
We empirically set $\lambda_{rec} = 1$, $\lambda_{perc} = 0.1$, $\lambda_{sty} = 250$, $\lambda_{G} = 0.1$, and $\lambda_{D} = 0.5$ in the total loss function (\eqnref{total_loss}).
For HorizonNet~\cite{SunHSC19}, we use the official pre-trained model for layout estimation.

\subsection{Evaluation on the Empty Scenes}
In this experiment, we evaluate both the qualitative and quantitative performance of our model on the image inpainting task by comparing it with baselines.
The qualitative comparisons are shown in~\figref{comp_sota_si_fr} (top 8 rows).
In contrast to EC and LISK, which fail to restore image structures in the masked regions, our method faithfully generates image contents adhering to the underlying layout structure.
While PanoDR shows slightly better structure preservation than EC and LISK, it fails to generate image contents consistent with the surrounding of masked regions as our method does.
Therefore, our method achieves the best performance against all the baselines across all evaluation metrics as shown in~\tabref{result_comp} (top).

\subsection{Evaluation on the Furnished Scenes}
Furniture of irregular shape will more or less obscure the layout of the indoor scene, making it more challenging to restore the regular structure in the missing area.
Therefore, in this experiment, we would like to evaluate how well our model learned from empty scenes can generalize to the furnished scenes.
Since the inpainting task setup here exactly matches the one defined in the PanoDR, we use the pre-trained model of PanoDR in this experiment for a fair comparison.
As shown in~\figref{comp_sota_si_fr} (bottom 8 rows), our method still clearly outperforms baselines in generating image contents that align the layout structure well and are consistent with the surrounding of the masked regions.
The quantitative results are shown in~\tabref{result_comp} (bottom).
It is worth noting that the way PanoDR performs image completion via compositing the predicted empty image and input image using the object mask will lead to severe artifacts where occlusion occurred between foreground objects (see~\figref{FR_SI}(PanoDR)).
\begin{table}[!t]
    \caption{\textbf{Quantitative comparisons with state-of-the-arts.} The top and bottom tables summarize the performance of our model and baselines on the empty and furnished scenes, respectively.}
    \label{tab:result_comp}
    \centering
        \begin{tabular}{l|l|llll}
        \toprule
        Dataset & Method & PSNR$\uparrow$ & SSIM$\uparrow$ & MAE$\downarrow$ & FID$\downarrow$           \\ 
        \midrule
        \multirow{4}{*}{Empty scene} & EC~\cite{nazeri2019edgeconnect} &38.6936 &0.9892 &0.0039	&3.9480 \\ 
                                    & LISK~\cite{jie2020inpainting}    &41.3761	&0.9895	&0.0055	&4.1660 \\ 
                                    & PanoDR~\cite{gkitsas2021panodr}  &37.2431	&0.9884	&0.0040	&4.3591 \\ 
                                    & Ours     &\tb{41.8444} &\tb{0.9919} &\tb{0.0030} &\tb{2.5265}     \\ 
        \midrule
        \multirow{4}{*}{Furnished scene} 
                                         & EC~\cite{nazeri2019edgeconnect} &31.4439 &0.9493	&0.0076 &11.9955 \\ 
                                         & LISK~\cite{jie2020inpainting}   &34.7325 &0.9553	&0.0068	&14.2676 \\
                                         & PanoDR~\cite{gkitsas2021panodr} &34.3340	&0.9641	&0.0051	&7.8399  \\ 
                                         & Ours     &\tb{35.3923} &\tb{0.9672} &\tb{0.0047} &\tb{7.2328}     \\ 
        \bottomrule
        \end{tabular}
\end{table}

\subsection{Ablation Study}
Here, we conduct ablation studies to validate our model from different perspectives.
First, we evaluate the necessity of individual design choices in our architecture.
Then, we conduct two experiments to evaluate how sensitive our model is to the size of input masks and the quality of input layout maps.
\begin{figure*}[t]
\centering
    \begin{subfigure}[t]{.19\textwidth}
       \includegraphics[width=\textwidth,keepaspectratio]{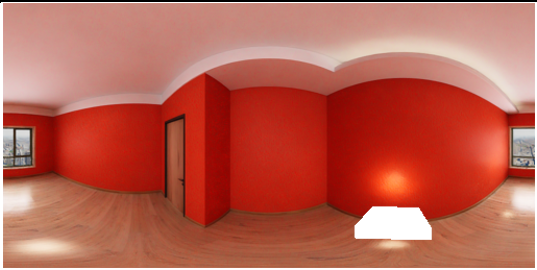}
       \includegraphics[width=\textwidth,keepaspectratio]{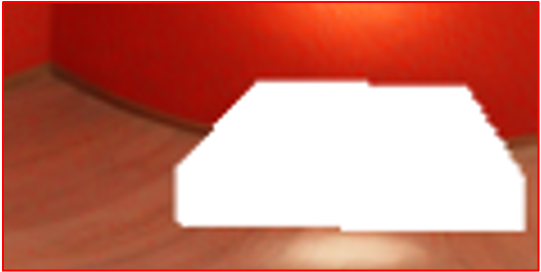}
       \includegraphics[width=\textwidth,keepaspectratio]{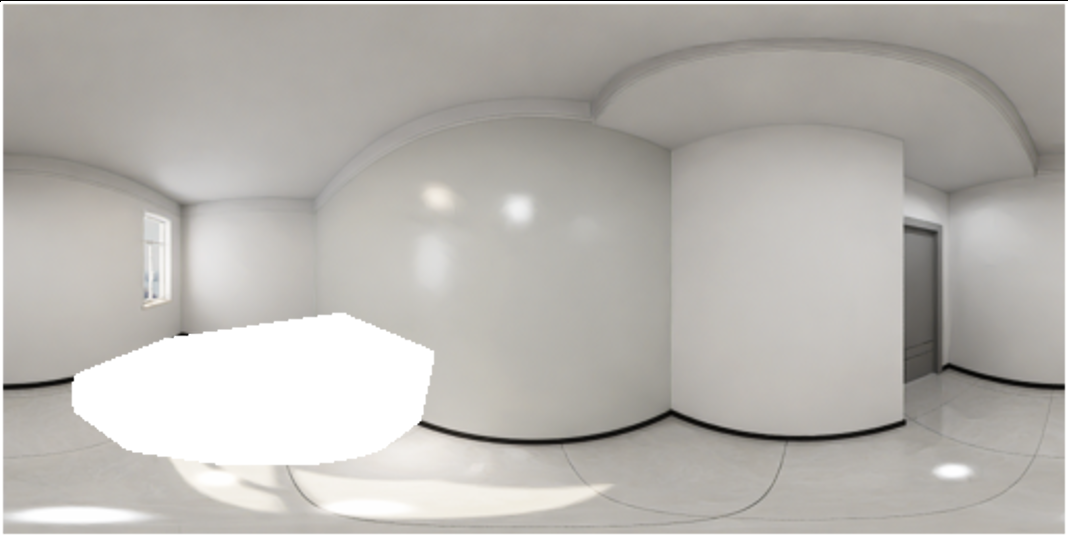}
       \includegraphics[width=\textwidth,keepaspectratio]{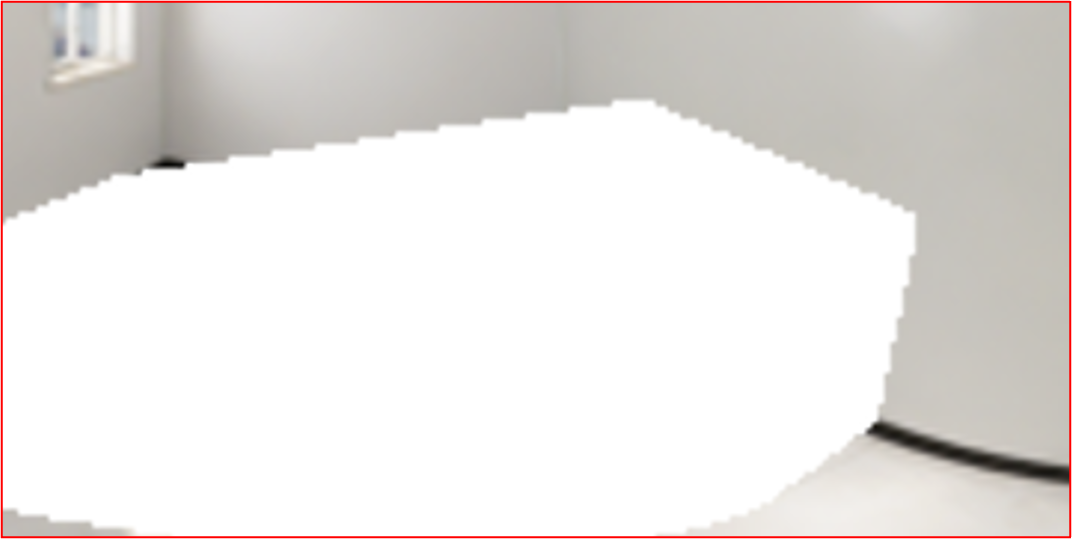}
       \includegraphics[width=\textwidth,keepaspectratio]{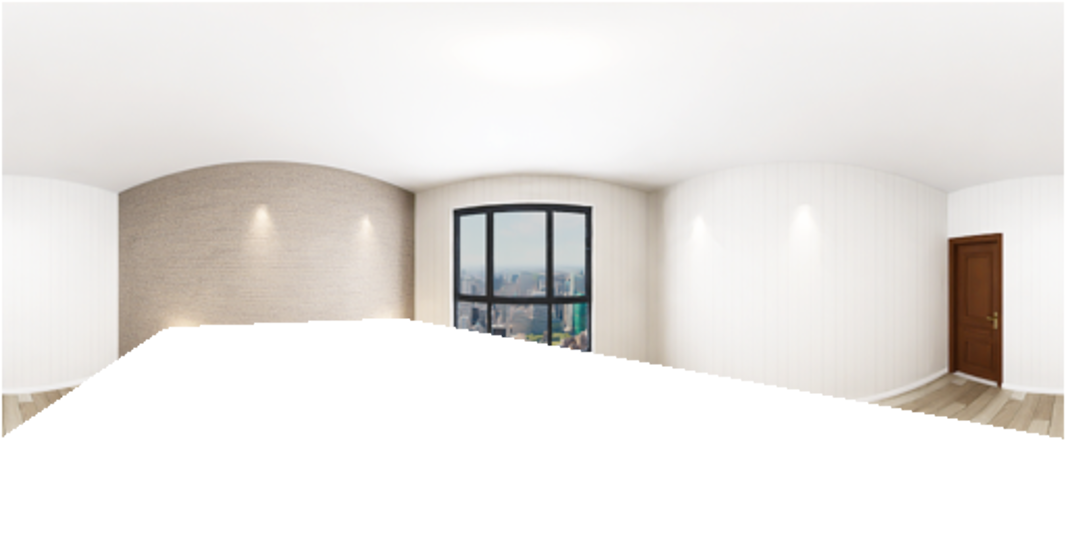}
       \includegraphics[width=\textwidth,keepaspectratio]{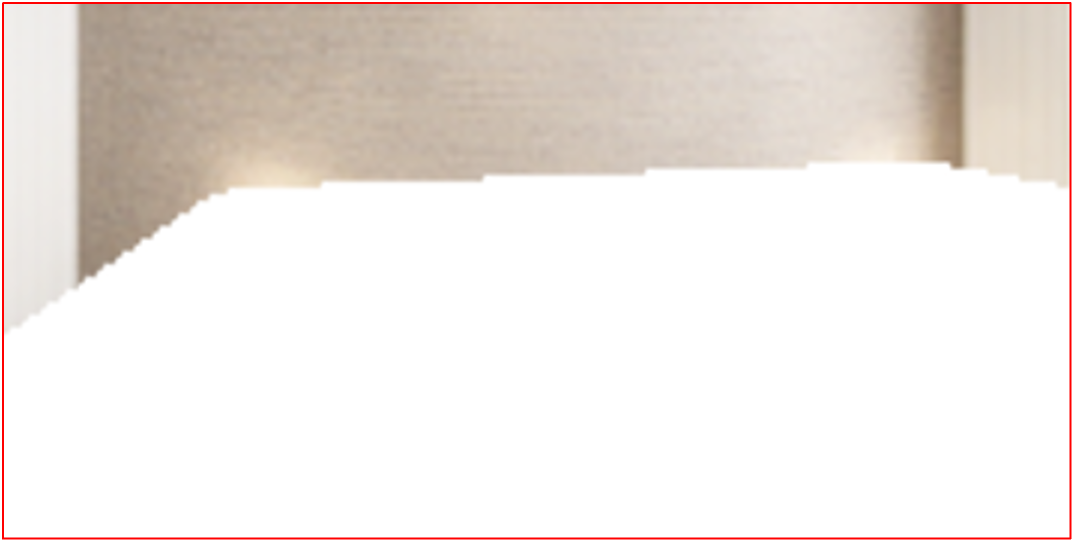}
       \caption*{Input}
    \end{subfigure}
    \begin{subfigure}[t]{.19\textwidth}
       \includegraphics[width=\textwidth,keepaspectratio]{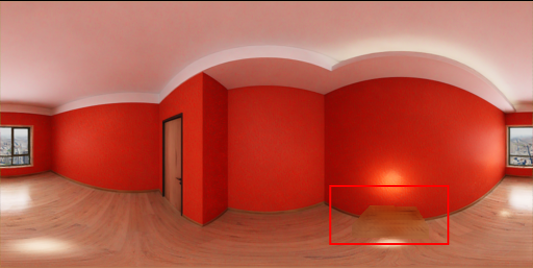}
       \includegraphics[width=\textwidth,keepaspectratio]{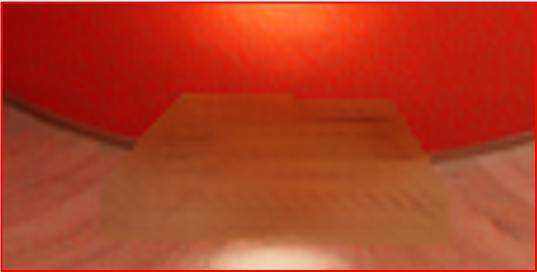}
       \includegraphics[width=\textwidth,keepaspectratio]{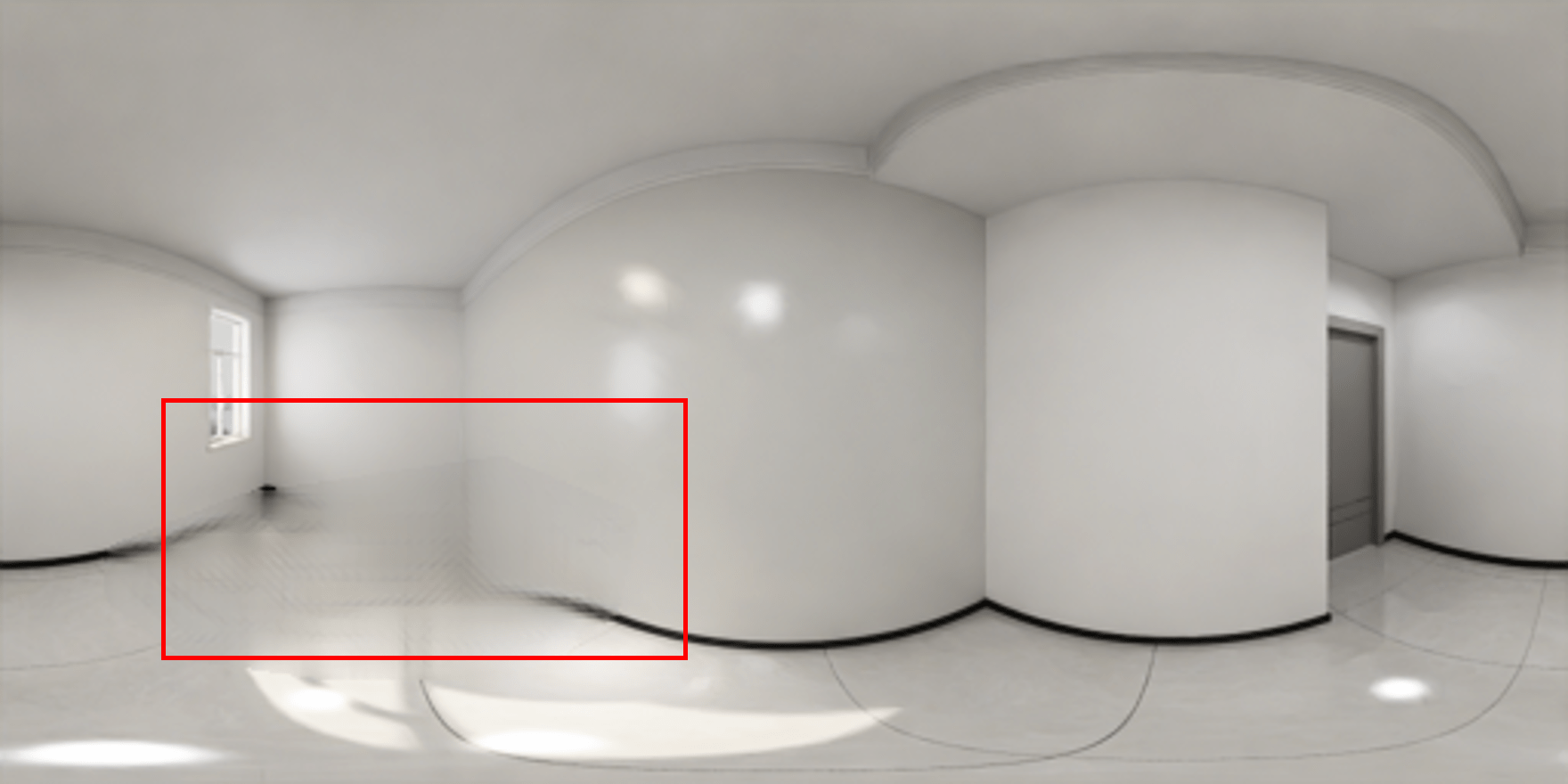}
       \includegraphics[width=\textwidth,keepaspectratio]{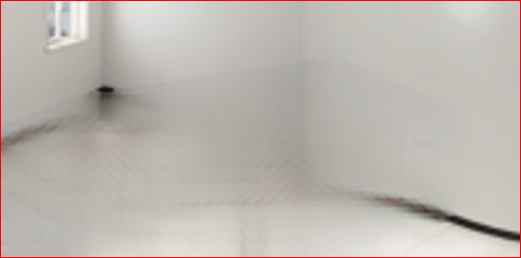}
       \includegraphics[width=\textwidth,keepaspectratio]{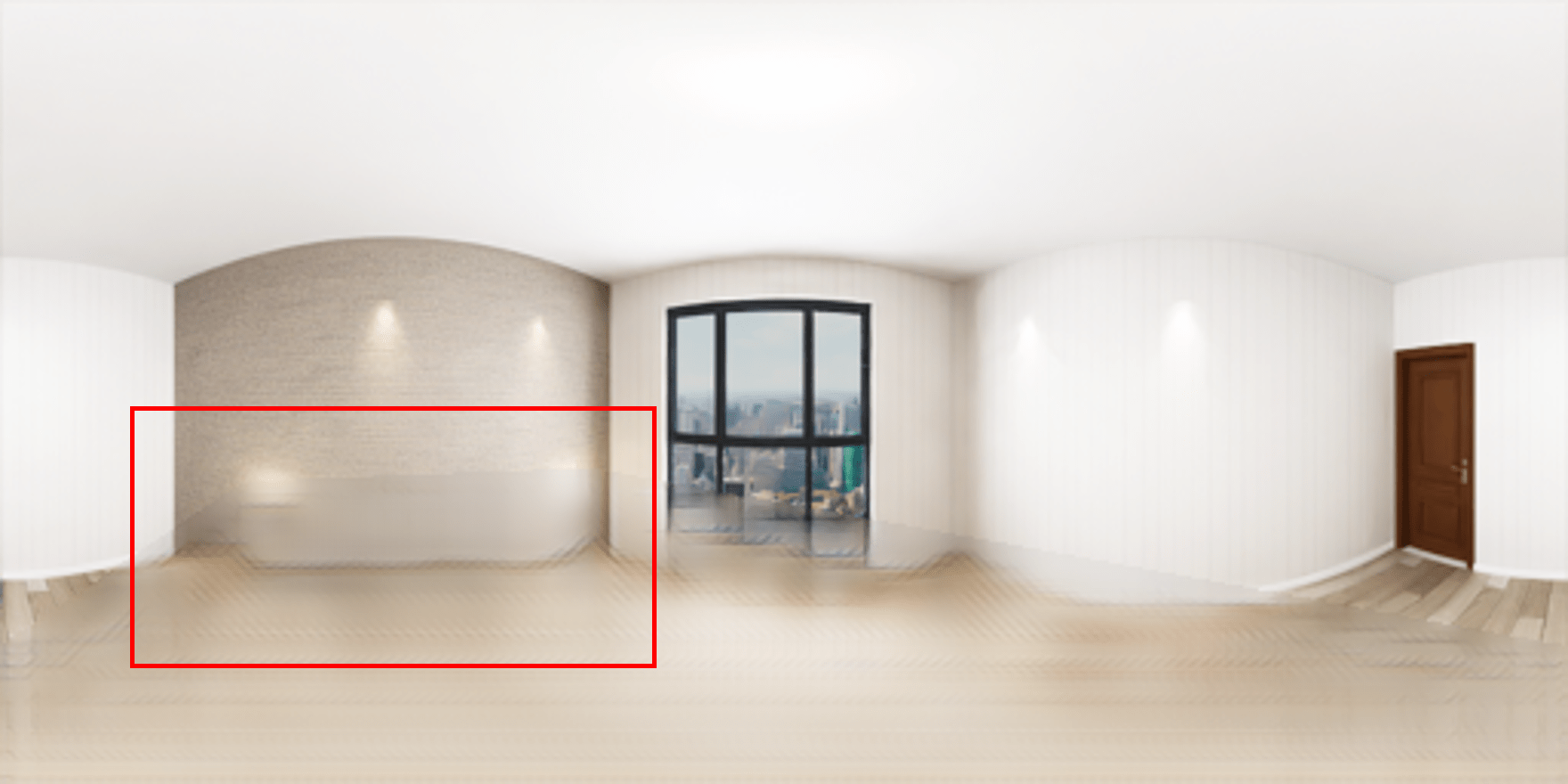}
       \includegraphics[width=\textwidth,keepaspectratio]{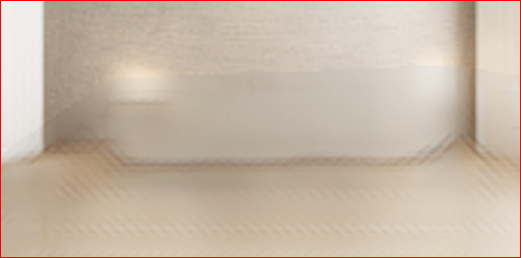}

       \caption*{\resizebox{.5\textwidth}{!}{$\baseline$}}
    \end{subfigure}
    \begin{subfigure}[t]{.19\textwidth}
       \includegraphics[width=\textwidth,keepaspectratio]{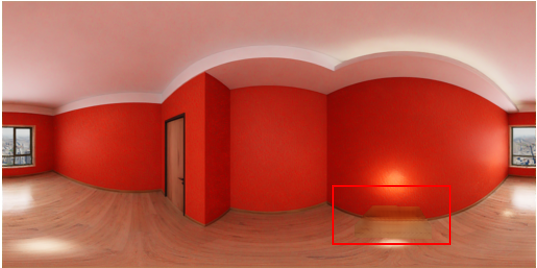}
       \includegraphics[width=\textwidth,keepaspectratio]{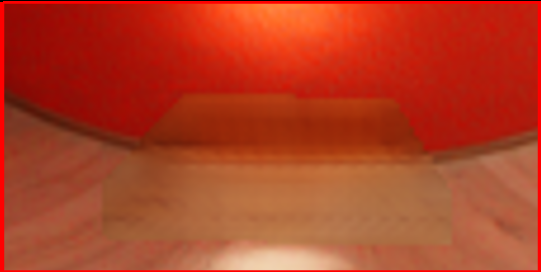}
       \includegraphics[width=\textwidth,keepaspectratio]{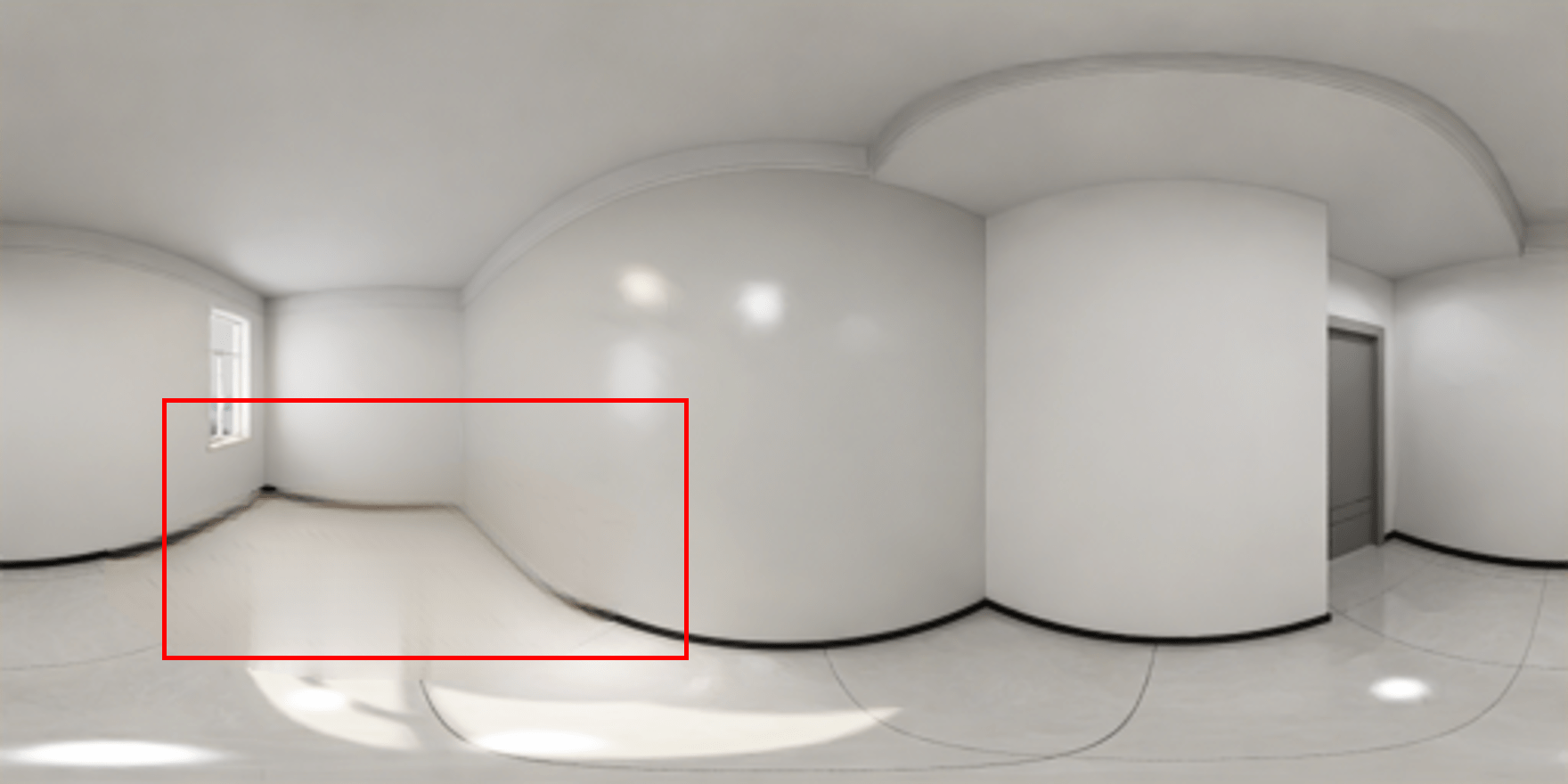}
       \includegraphics[width=\textwidth,keepaspectratio]{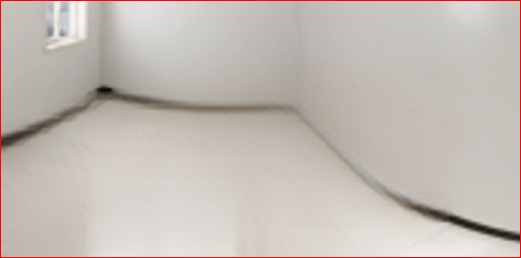}
       \includegraphics[width=\textwidth,keepaspectratio]{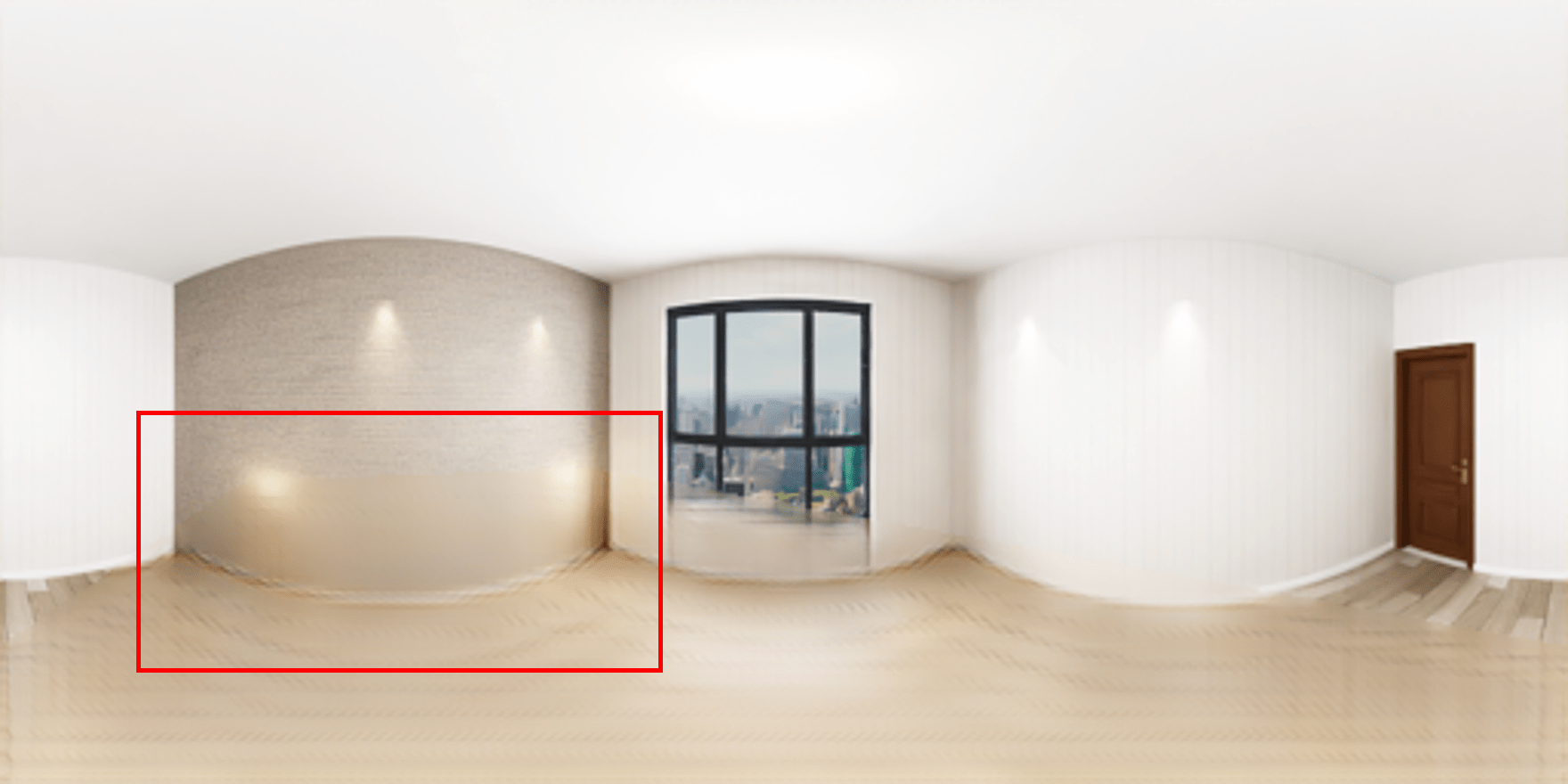}
       \includegraphics[width=\textwidth,keepaspectratio]{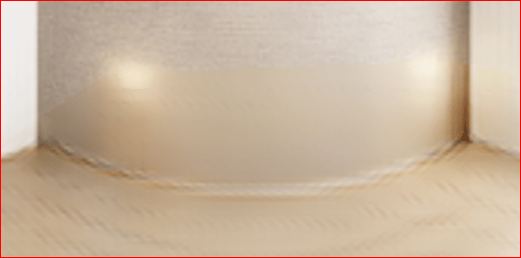}

       \caption*{\resizebox{.85\textwidth}{!}{$\layout$}}
    \end{subfigure}
    \begin{subfigure}[t]{.19\textwidth}
       \includegraphics[width=\textwidth,keepaspectratio]{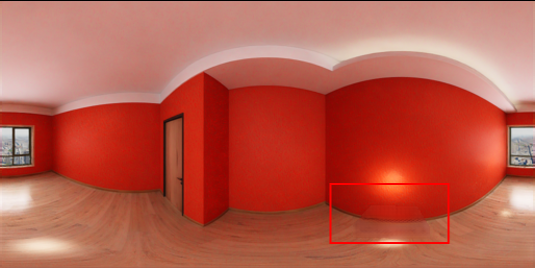}
       \includegraphics[width=\textwidth,keepaspectratio]{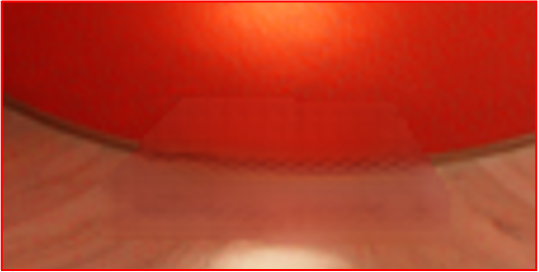}
       \includegraphics[width=\textwidth,keepaspectratio]{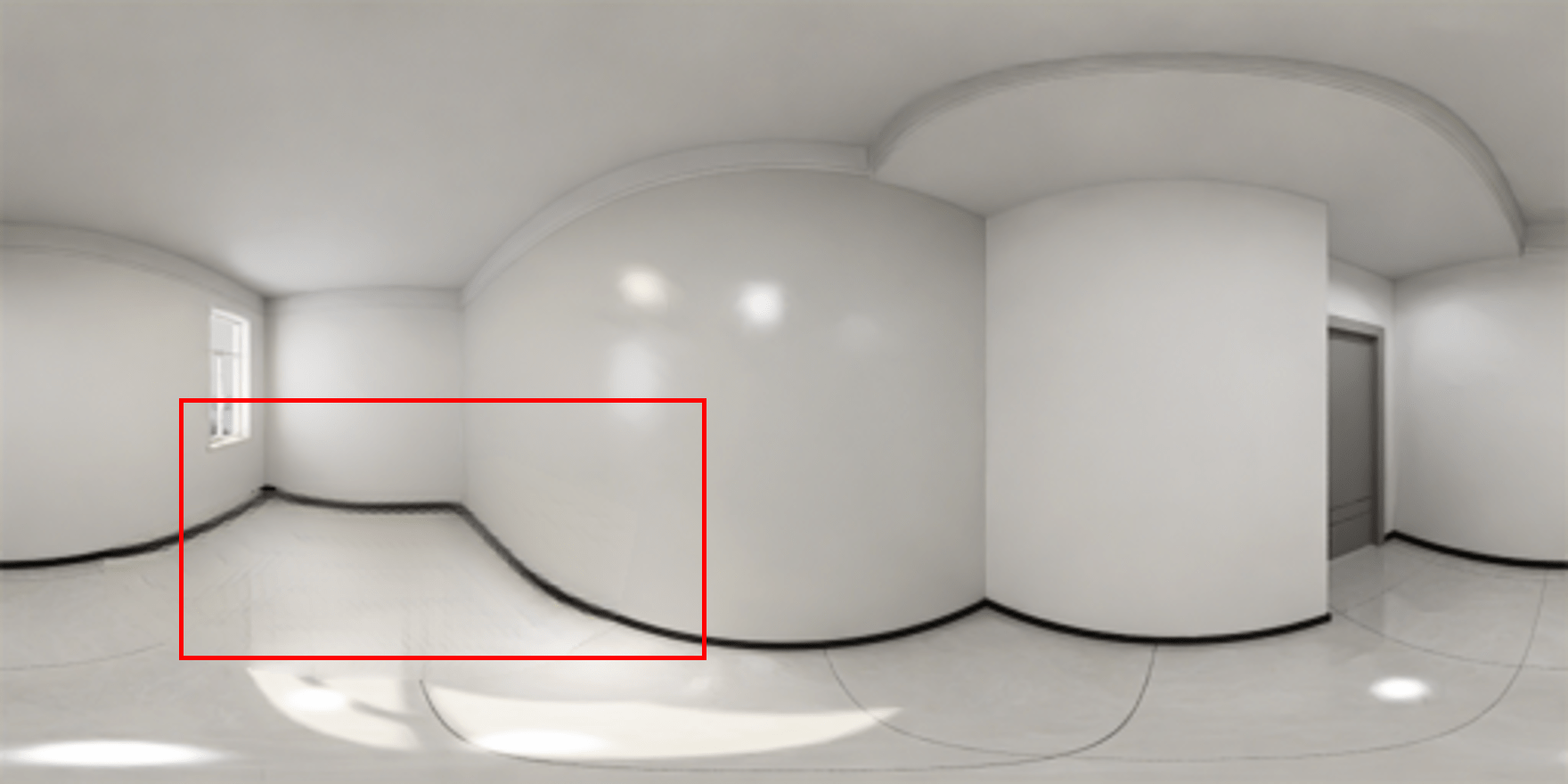}
       \includegraphics[width=\textwidth,keepaspectratio]{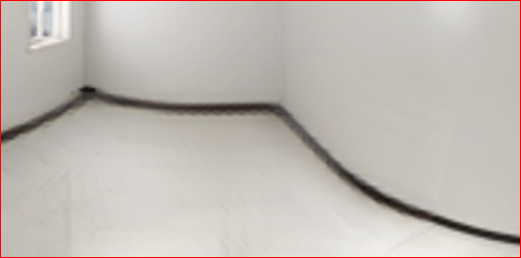}
       \includegraphics[width=\textwidth,keepaspectratio]{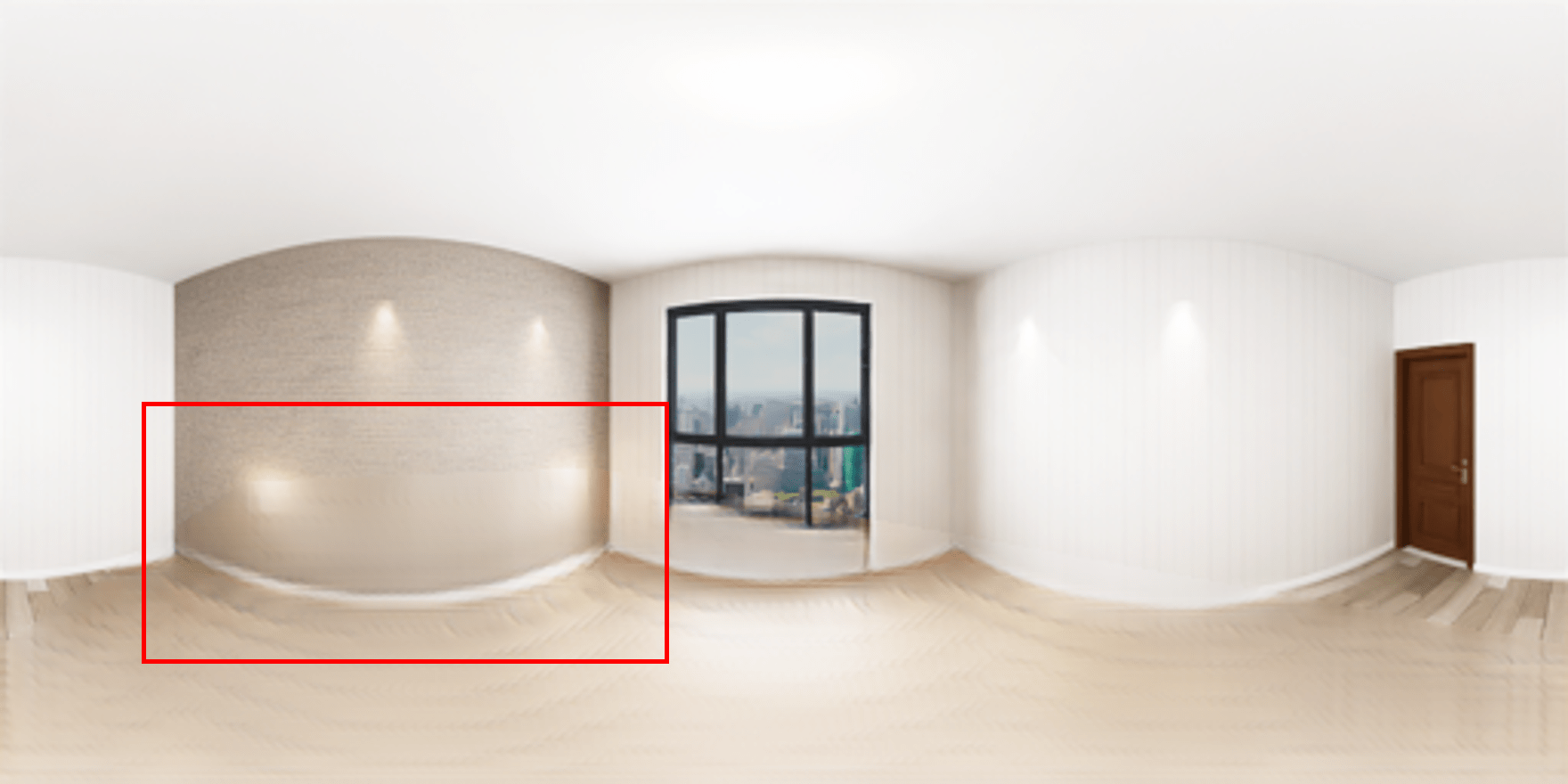}
       \includegraphics[width=\textwidth,keepaspectratio]{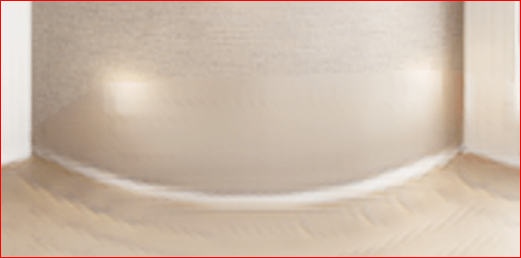}

       \caption*{\resizebox{.55\textwidth}{!}{$\planarawarenor$}}
    \end{subfigure}
    \begin{subfigure}[t]{.19\textwidth}
       \includegraphics[width=\textwidth,keepaspectratio]{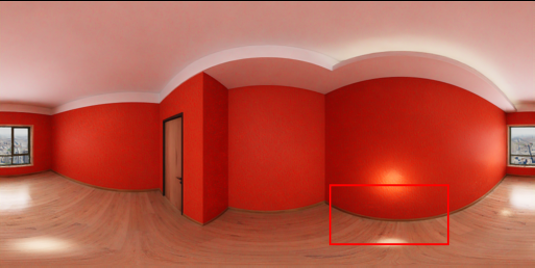}
       \includegraphics[width=\textwidth,keepaspectratio]{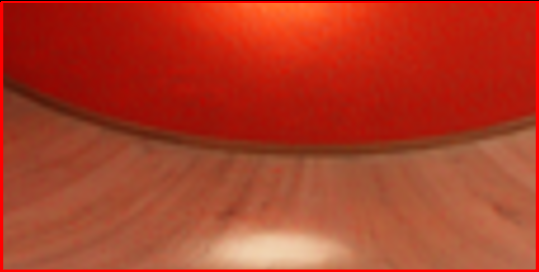}
       \includegraphics[width=\textwidth,keepaspectratio]{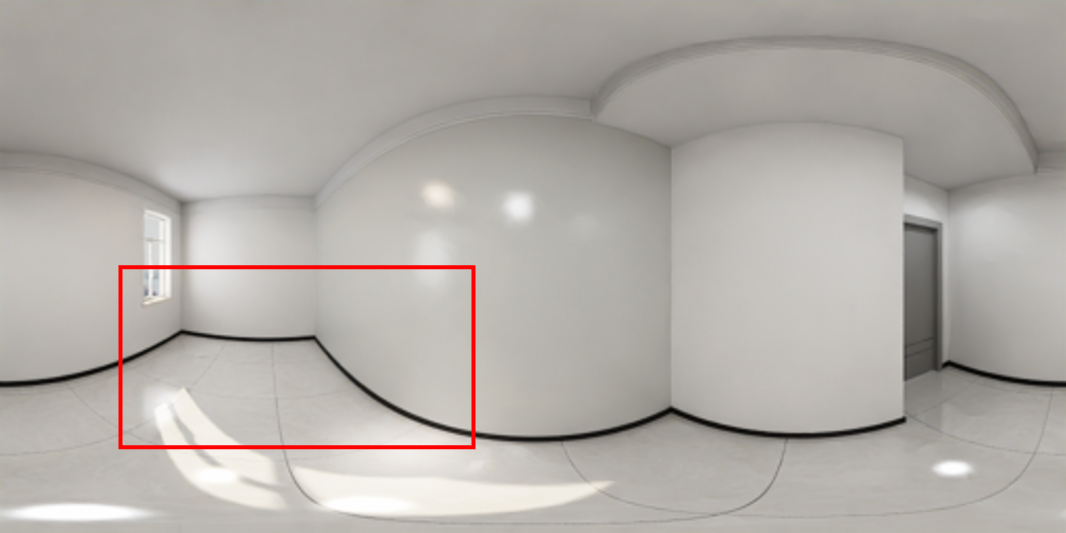}
       \includegraphics[width=\textwidth,keepaspectratio]{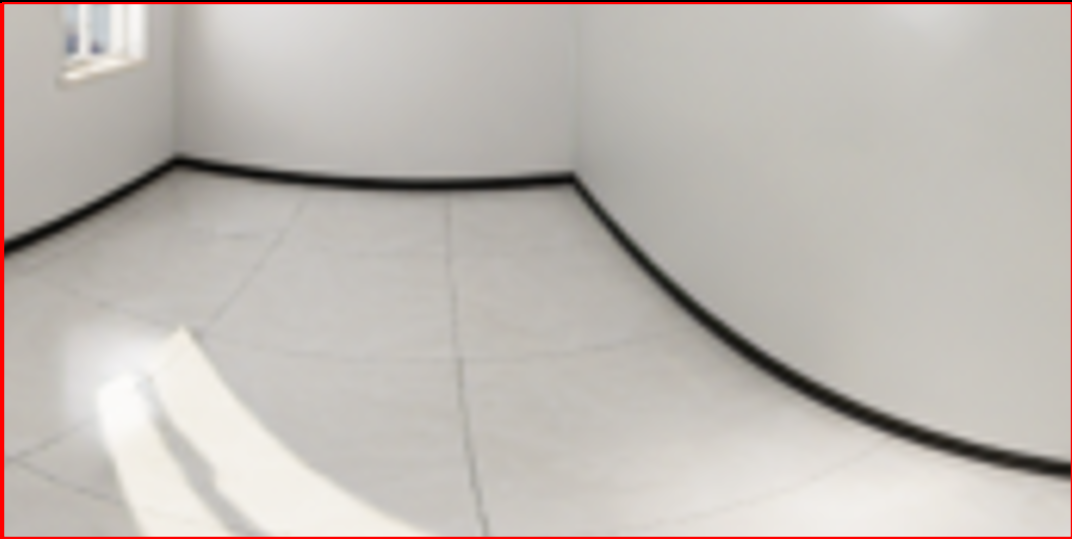}
       \includegraphics[width=\textwidth,keepaspectratio]{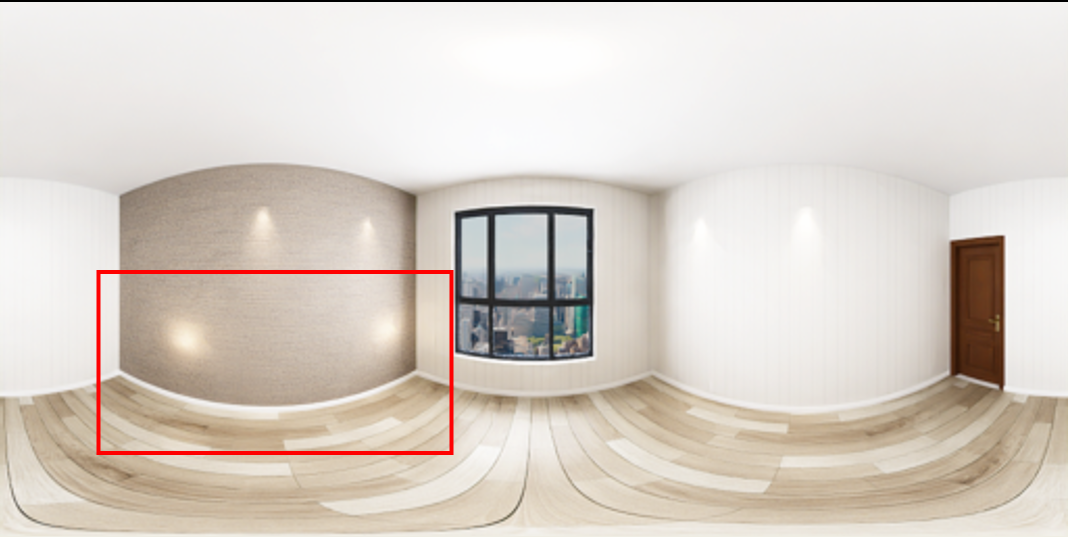}
       \includegraphics[width=\textwidth,keepaspectratio]{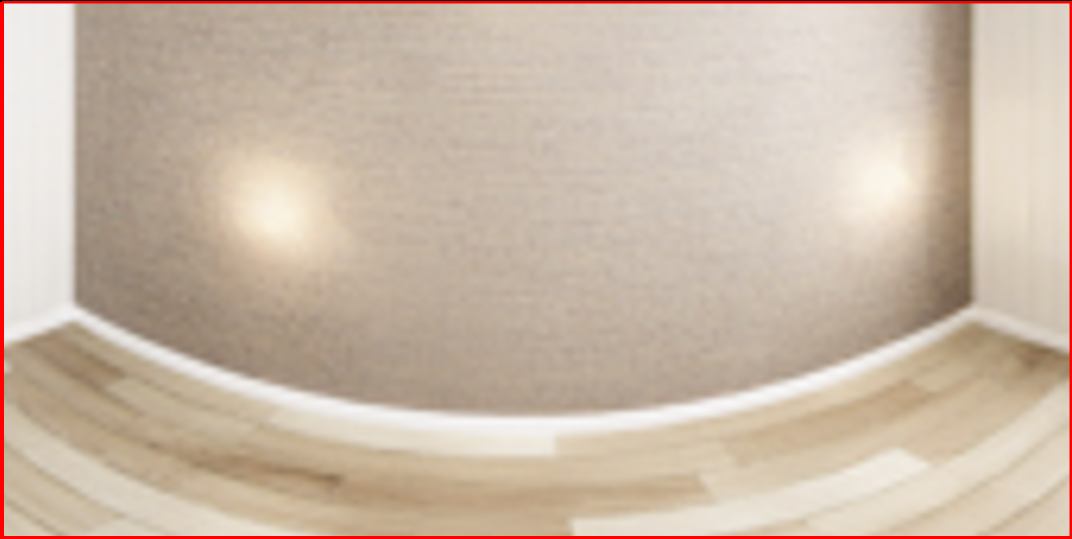}

       \caption*{GT}
    \end{subfigure}
    \vspace{\figcapmargin}
    \caption{\textbf{Qualitative results of the ablation study.} Side-by-side comparisons of inpainting results generated using our method by gradually adding individual components. From left to right, input images and masks, our baseline model ($\baseline$), adding the layout guidance map ($\layout$), full model with our plane-aware normalization ($\planarawarenor$), and ground truth images.}
    \label{fig:ablation_new}
\end{figure*}
\begin{table}[!t]
    \caption{\textbf{Quantitative results of the ablation study.} We evaluate the effectiveness of our design choices by gradually adding the individual components into the architecture.}
    \label{tab:ablation_study}
    \centering
        \begin{tabular}{lllllll}
        \toprule
        \phantom{aaa} & PSNR $\uparrow$ & SSIM $\uparrow$ & MAE $\downarrow$  & FID $\downarrow$\\
        \midrule
        $\baseline$   &40.6449	    &0.9911	        &0.0034	        &3.3915 \\
        $\layout$     &41.2884	    &0.9916	        &0.0033	        &2.8105\\
        $\planarawarenor$  &\tb{41.8444}	&\tb{0.9919}    &\tb{0.0030}	&\tb{2.5265} \\ 
        \bottomrule
        \end{tabular}
\end{table}

\heading{Ablation on network architecture.}
In this experiment, we start with the backbone model ($\baseline$) as the baseline, then progressively adding only layout guidance map ($\layout$), and our plane-aware normalization ($\planarawarenor$).
As shown in~\tabref{ablation_study}, we obtain the best performance with the full model on all the metrics.
The qualitative comparisons shown in~\figref{ablation_new} indicate that adding layout guidance map generates clear structure boundaries in the final result ($2^{nd}$ and $3^{rd}$ columns), while our full model with plane-aware normalization can constrain the image generation to the adjacent structural planes and obtain visually consistent results ($3^{rd}$ and $4^{th}$ columns).

\heading{Sensitivity to the mask size.}
In this experiment, we analyze the testing dataset and classify the images into different categories according to the area proportions of input masks. \tabref{add_exp1} shows the inpainting performance for each category. We can tell that the inpainting quality degrades with the increasing mask size. A significant drop occurs where the ratio of input mask is greater than 30{\%}.  

\heading{Sensitivity to the layout estimation.}
In order to explore the effect of the accuracy of layout estimation on the inpainting quality, we first devise a mechanism to generate layout maps with different levels of accuracy. 
Specifically, we feed masked images of different mask sizes into HorizonNet. We start by generating randomly located rectangle masks of 5\% image size and increase the mask ratio to 10\%, 30\% and 50\% to deliberately produce layout structures with decreasing quality.
Then we take these layout maps as conditional inputs of our model and compare the inpainting performance empty-room testing dataset.
As shown in~\tabref{add_exp2}, our model degrades marginally when the quality of estimated layouts decreases from 0.96 to 0.84, indicating our model is robust to the varying input layout maps.
\begin{table}[!t]
    \caption{\textbf{Mask size vs. inpainting quality.}}
    \label{tab:add_exp1}
    \centering
        \begin{tabular}{lllll}
        \toprule
        \multirow{2}{*}{Mask Size(\%)} &
        \multirow{2}{*}{Count} \phantom{ab} &
        \multicolumn{3}{c}{Content} \\
        \cmidrule{3-5}
         {} & & PSNR $\uparrow$ & SSIM $\uparrow$ & MAE $\downarrow$ \\
        \midrule
        0-10    &1045	    &44.3921 	    &0.9967	    &0.0011     \\
        10-20    &163	    &34.2823 	    &0.9841	    &0.0055     \\
        20-30    &48 	    &30.4371 	    &0.9726	    &0.0111     \\
        30-40    &39 	    &25.0731 	    &0.9386	    &0.0266     \\
        40+     &13 	    &24.2958 	    &0.9345	    &0.0305     \\
        total     &1308 	    &41.8444 	    &0.9919	    &0.0030     \\
        \bottomrule
        \end{tabular}
\end{table}

\begin{table}[!t]
    \caption{\textbf{Accuracy of layout estimation vs. inpainting quality.}}
    \label{tab:add_exp2}
    \centering
        \begin{tabular}{lllll}
        \toprule
        \multicolumn{1}{l}{Structure} \phantom{ab} & \multicolumn{4}{c}{Content} \\
        \cmidrule{1-5}
         mIOU $\uparrow$ & PSNR $\uparrow$ & SSIM $\uparrow$ & MAE $\downarrow$ & FID $\downarrow$  \\
        \midrule
            0.9603	    &42.3212 	    &0.9925	    &0.0028     &2.4322	    \\
            0.9561	    &42.2871 	    &0.9925	    &0.0028     &2.4441     \\
            0.9175 	    &42.0682 	    &0.9923	    &0.0029     &2.5624	    \\
            0.8489 	    &41.7300 	    &0.9919	    &0.0030     &2.8455	    \\
        \bottomrule
        \end{tabular}
\end{table}
\begin{figure*}[t]
\centering
    \begin{subfigure}[t]{.19\textwidth}
    \includegraphics[width=\textwidth,keepaspectratio]{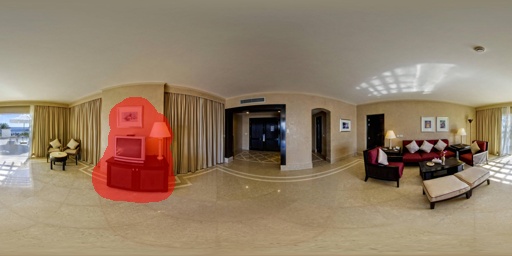}
       
       \includegraphics[width=\textwidth,keepaspectratio]{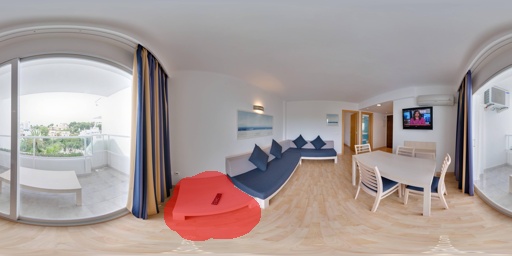}
       \includegraphics[width=\textwidth,keepaspectratio]{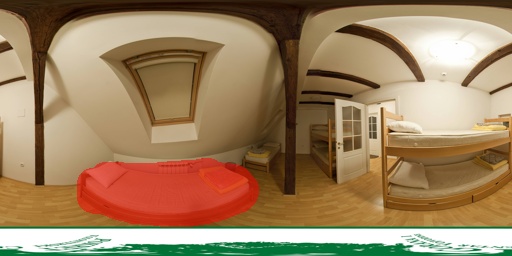}

       \caption*{Input}
    \end{subfigure}
    \begin{subfigure}[t]{.19\textwidth}
    \includegraphics[width=\textwidth,keepaspectratio]{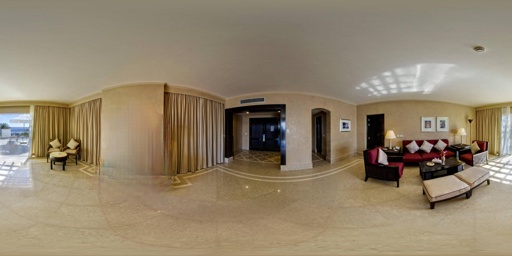}
       \includegraphics[width=\textwidth,keepaspectratio]{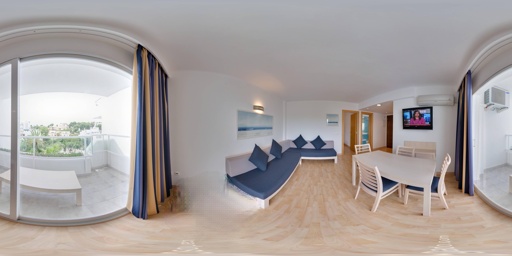}
       \includegraphics[width=\textwidth,keepaspectratio]{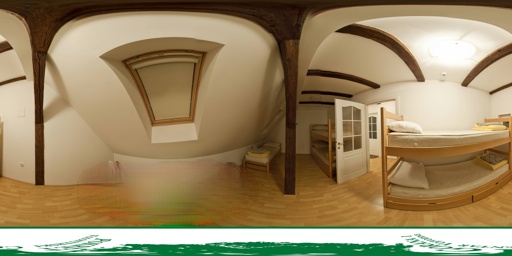}

       \caption*{EC~\cite{nazeri2019edgeconnect}}
    \end{subfigure}
    \begin{subfigure}[t]{.19\textwidth}
    \includegraphics[width=\textwidth,keepaspectratio]{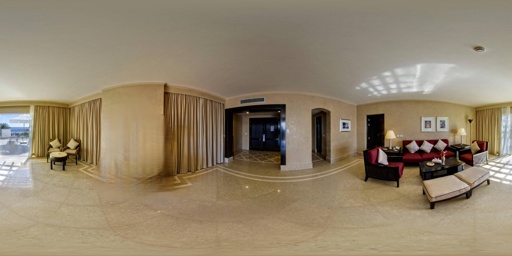}
       \includegraphics[width=\textwidth,keepaspectratio]{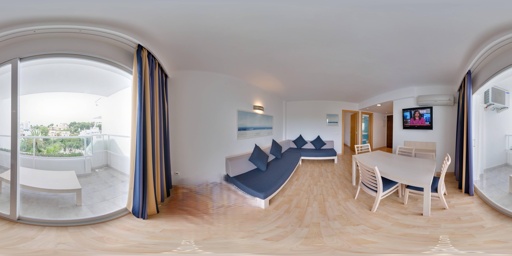}
       \includegraphics[width=\textwidth,keepaspectratio]{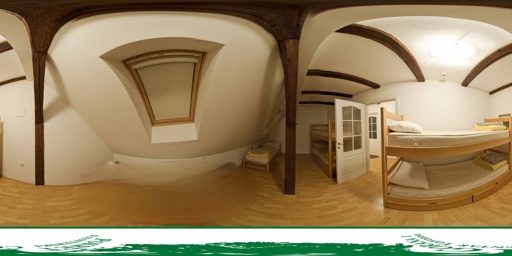}

       \caption*{LISK~\cite{jie2020inpainting}}
    \end{subfigure}
    \begin{subfigure}[t]{.19\textwidth}
    \includegraphics[width=\textwidth,keepaspectratio]{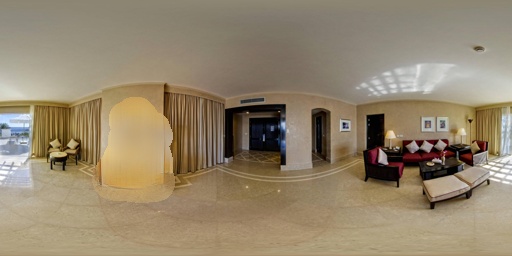}
       \includegraphics[width=\textwidth,keepaspectratio]{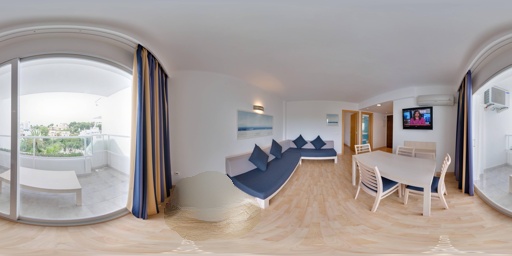}
       \includegraphics[width=\textwidth,keepaspectratio]{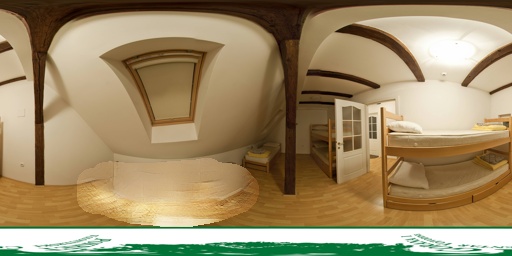}

       \caption*{PanoDR~\cite{gkitsas2021panodr}}
    \end{subfigure}
    \begin{subfigure}[t]{.19\textwidth}
    \includegraphics[width=\textwidth,keepaspectratio]{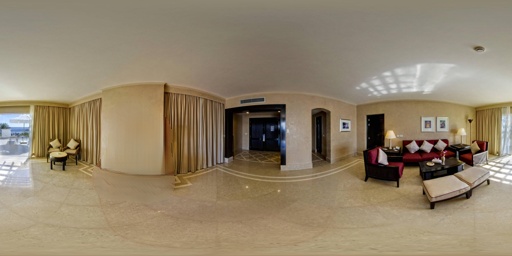}
       \includegraphics[width=\textwidth,keepaspectratio]{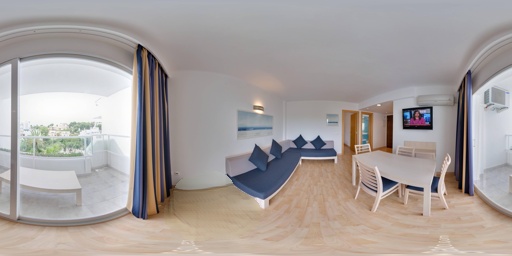}
       \includegraphics[width=\textwidth,keepaspectratio]{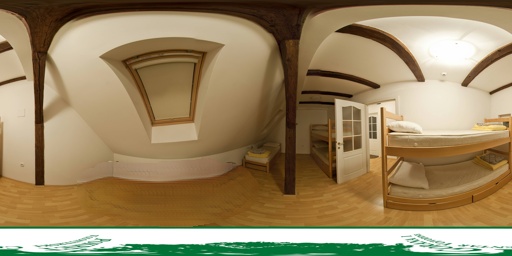}

       \caption*{Ours}
    \end{subfigure}
    \caption{\textbf{Qualitative comparisons with state-of-the-arts on real-world scenes.} Our model clearly outperforms baselines by preserving layout boundary and restoring local texture from adjacent room structures (\ie floor and walls).
    }
    \label{fig:real_scene}
\end{figure*}

\subsection{Qualitative Results on Real-world Scene}
Real-world scenes have complex lighting and layout structure.
However, the amount of data in the real-world scene dataset and the quality of furniture category annotations are insufficient for training our model, so we choose to train on the synthetic dataset Structured3D~\cite{Structured3D}.
Nevertheless, we still compare our results with PanoDR~\cite{gkitsas2021panodr}, which also implements the furniture removal task, on the real-world scene dataset. 
Since the real-world scene dataset does not contain paired data (\ie scenes before and after furniture removal), quantitative evaluation is infeasible and we can only provide qualitative comparisons here.
\figref{real_scene} shows that our inpainted results have a higher quality of structural maintenance and color restoration. Moreover, compared with PanoDR, we can still exert more stable performance in real-world scenes.
Please refer to our online webpage for more results\footnote{\url{https://ericsujw.github.io/LGPN-net/}}.
\section{Conclusions}
\label{sec:conclusions}
We proposed an end-to-end structural inpainting network for the indoor scene. 
We introduce layout boundary line conditions the output structure and utilize the plane-aware normalization to enhance planar style consistency. 
Experiment results show the outstanding performance of our model in both structural inpainting and furniture removal on the indoor scene.

\begin{figure}[!t]
\centering
    \begin{subfigure}[t]{.32\columnwidth}
       \includegraphics[width=\columnwidth,keepaspectratio]{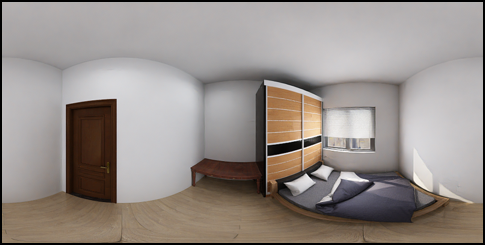}
       \includegraphics[width=\columnwidth,keepaspectratio]{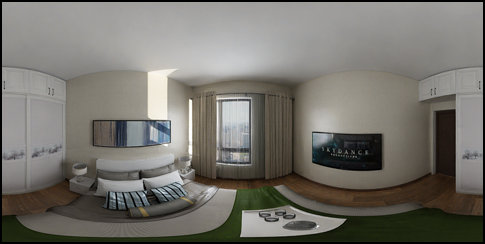}

       \caption*{Input}
    \end{subfigure}
    \begin{subfigure}[t]{.32\columnwidth}
       \includegraphics[width=\columnwidth,keepaspectratio]{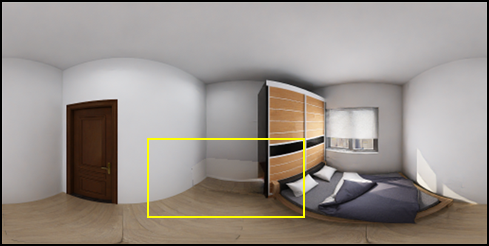}
       \includegraphics[width=\columnwidth,keepaspectratio]{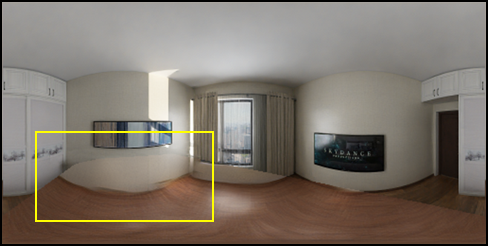}
       \caption*{PanoDR~\cite{gkitsas2021panodr}}
    \end{subfigure}
    \begin{subfigure}[t]{.32\columnwidth}
       \includegraphics[width=\columnwidth,keepaspectratio]{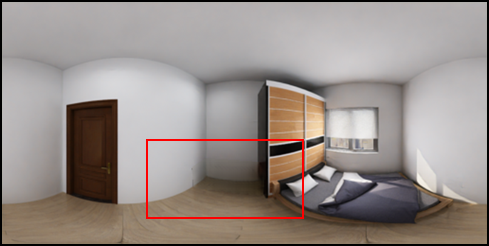}
       \includegraphics[width=\columnwidth,keepaspectratio]{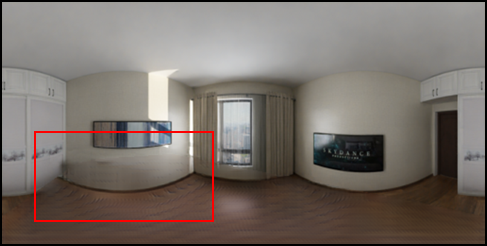}
       \caption*{Ours}
    \end{subfigure}
    \caption{\textbf{Limitation.} Both the state-of-the-art method and our model produce visual artifacts in the scenes presenting strong shading effect surrounding the removed furniture.}
    \label{fig:Limitation}
\end{figure}

\heading{Limitations.} 
In the real-world application of furniture removal, we can often see residuals of shading effect caused by the removed furniture. 
These residuals are hard to segment and even harder to model.
As shown in~\figref{Limitation}, our model is slightly affected by these residuals but still produces more realistic results than PanoDR~\cite{gkitsas2021panodr}.

\heading{Future work.}
We plan to adopt a more reasonable segmentation mask of the indoor scene inpainting which can cover the shading area and thus improve our results in those shaded scenes.

\heading{Acknowledgements.} The project was funded in part by the National Science and Technology Council of Taiwan (110-2221-E-007-060-MY3, 110-2221-E-007-061-MY3).
\bibliographystyle{splncs04}
\bibliography{egbib}

\end{document}